%% file: main.tex
\documentclass{article}

\usepackage{microtype}
\usepackage{graphicx}
\usepackage{subfigure}
\usepackage{booktabs} 
\usepackage{array}
\usepackage{hyperref}

\newcolumntype{C}{>{\centering\arraybackslash}m{6em}}

 \usepackage[accepted]{icml2023}

\usepackage{amsmath}
\usepackage{amssymb}
\usepackage{mathtools}
\usepackage{amsthm}

\usepackage[capitalize,noabbrev]{cleveref}

\usepackage[textsize=tiny]{todonotes}

\input{math_commands.tex}

\input{preamble}
\usepackage{caption}
\usepackage{amssymb,amsfonts,amsmath,amsthm,amscd,dsfont,mathrsfs,bbm}
\usepackage{graphicx,float,psfrag,epsfig,amssymb}
\definecolor{darkgreen}{rgb}{0.0,0,0.9}
\usepackage{wrapfig}
\usepackage{relsize}
\usepackage{color}
\usepackage{pict2e}
\usepackage{nameref}
\usepackage{makecell}
\usepackage{mathtools}
\usepackage{enumitem}
\usepackage{subfigure}

\usepackage{algorithm}

\newtheorem{propo}{Proposition}[section]

\newtheorem{proposition}[propo]{Proposition}
\newtheorem{defi}[propo]{Definition}

\newtheorem{thm}[propo]{Theorem}

\usepackage{hyperref}       
\usepackage{url}   

\newcommand{\indep}{\perp \!\!\! \perp}

 \input{notation}

\def\tz{\widetilde{Z}}

\def\cA{\mathcal{A}}

\def\normal{\mathsf{N}}
\def\tv{\mathsf{TV}}
\def\kl{\mathsf{KL}}

\def\bX{\mathbf{X}}
\def\bY{\mathbf{Y}}
\def\bx{\mathbf{x}}
\def\by{\mathbf{y}}
\def\tbx{\widetilde{\mathbf{x}}}
\def\tby{\widetilde{\mathbf{y}}}

\def \swap {\mathsf{swap}}
\def \trace {\mathsf{tr}}
\def\tswap{\tau_X}

\icmltitlerunning{ A Model-free Closeness-of-influence Test for Features in Supervised Learning
}

\begin{document}

\twocolumn[
\icmltitle{A Model-free Closeness-of-influence Test for Features in Supervised Learning}



\begin{icmlauthorlist}
\icmlauthor{Mohammad Mehrabi}{sch}
\icmlauthor{Ryan A. Rossi}{comp}
\end{icmlauthorlist}

\icmlaffiliation{sch}{Department of Data Sciences and Operations, University of Southern California, Los Angeles, USA}
\icmlaffiliation{comp}{Adobe, San Jose, USA}

\icmlcorrespondingauthor{Mohammad  Mehrabi}{mehrabim@marshall.usc.edu}

\icmlkeywords{feature influence, nonparametric testing}

\vskip 0.3in
]



\printAffiliationsAndNotice{}
\begin{abstract}
Understanding the effect of a feature vector $x\in \reals^d$  on the response value (label) $y\in \reals$ is the cornerstone of many statistical learning problems. Ideally, it is desired to understand how a set of collected features combine together and influence the response value, but this problem is notoriously difficult, due to the high-dimensionality of data and limited number of labeled data points, among many others. 
%
In this work, we take a new perspective on this problem, and  we study the question of assessing the difference of influence that the two given features have on the response value. 
We first propose a notion of closeness for the influence of features, and show that our definition recovers the familiar notion of the magnitude of coefficients in the parametric model. We then propose a novel method to test for the closeness of influence  in  general model-free supervised learning problems. 
 Our proposed test can be used with finite number of samples with control on type I error rate, no matter the ground truth conditional law $\mathcal{L}(Y|X)$. We analyze the power of our test for two general learning problems i) linear regression, and ii) binary classification under mixture of Gaussian models, and show that under the proper choice of score function, an internal component of our test, with sufficient number of samples will achieve full statistical power. We evaluate our findings through extensive numerical simulations, specifically we adopt the datamodel framework (Ilyas, et al., 2022) for CIFAR-10 dataset to identify pairs of training samples with different influence on the trained model via optional black box training mechanisms.  
\end{abstract}

\section{Introduction}

In a classic supervised learning problem, we are given a dataset of $n$ iid data points $\{(x_i,y_i)\}_{i=1:n}$ with feature vectors $x\in \reals^d$ and response value (label) $y\in \reals$. From the inferential point of view, understanding the influence of each individual feature $i\in \{1,\dots,d\}$ on $y$ is of paramount importance. Considering a parametric family of distributions for $\cL(Y|X)$ is among the most studied techniques for this problem. In this setting, the influence of each feature can be seen by their corresponding coefficient value in the parametric model. Essentially such methods can result in spurious statistical findings, mainly due to model misspecification, where in the first place the ground-truth data generating law $\cL(Y|X)$ does not belong to the considered parametric family. A natural remedy for this problem is to relax the parametric family assumption, removing concerns about model misspecification. Besides the difficulties with the new model-free structure of the problem, we need a new notion to capture the influence of features, as there is no longer a coefficient vector as per class of parametric models.  



In this paper, we follow the model-free structure, but take a new perspective on the generic problem of investigating the influence of features on the response value. In particular, as a first step towards this notoriously hard question under  no class of parametric distribution assumption or whatsoever, we are specifically interested in assessing the closeness of influence of features.  For this end,  we posit the following fundamental question:
\begin{center}
    \textit{ (*) In a general model-free supervised learning problem, for two given features, is it possible to assess the closeness of their influence on the response value (label) in a statistically sound way?}
\end{center}


In this paper, we answer question (*) affirmatively. We characterize a notion of closeness for the influence of features on $y$ under the general model-free framework. We show that this notion aligns perfectly well with former expectations in parametric models, where small difference in the coefficient values imply close influence on the response value. We then cast the closeness of influence question as a hypothesis testing problem, and show that we can control associated type I error rate with finite number of samples. 



\subsection{Motivation Behind Question (*)}
Beyond the inferential nature of Question (*) that
helps to better understand the data-generating process of on-hand data, being able to answer this question has a myriad of applications for other classic machine learning tasks.  In fact, inspired by the recent advancements in interpretable machine learning systems, it is desired to strike a balance between model flexibility in capturing the ground-truth law $\mathcal{L}(Y|X)$ and using few number of explanatory variables. For this goal, feature aggregation has been used to distill a large amount of feature information into a smaller number of features. In several parametric settings, features with equal coefficients are naturally grouped together, e.g, in linear regression new feature $x_1+x_2$ is considered rather than $(x_1,x_2)$, in case that $x_1,x_2$ have equal corresponding regression coefficients \citep{yan2021rare}. In addition, identifying features with near influence on the response value can be used for tree-based aggregation schemes~\citep{shao2021controlling,bien2021tree, wilms2022tree}. This is of paramount importance in learning problems involving rare features, such as the count of microbial species \citep{bien2021tree}.  In addition, in many learning problems, an honest comprehensive assessment for characterizing the behavior of $Y$ with respect to a certain attribute $A$ is desired. This can be used to assess the performance of model with respect to a sensitive attribute (fair machine learning), or to check if two different treatments (different values of $A$) have close influence on potential outcomes.  

\subsection{Related Work}
In machine learning, the problem of identifying a group of features that have the largest influence on the response value is often formulated as \text{variable selection}.
With a strong parametric assumption, the conditional law $\cL(Y|X)$ is considered to belong to a known class of parametric models, such as linear regression. For variable selection in the linear regression setting, the LASSO \citep{tibshirani1996regression} and Dantzig selector \citep{candes2007dantzig} are the most widely used. In fact, there are several other works for variable selection in the linear regression setting with output solutions satisfying certain structures, such as \citep{bogdan2015slope, tibshirani2005sparsity}. 
There has been another complimentary line in the past years from model-X perspective \citep{candes2018panning}. In this setting, despite the classical setup, in which a strong parametric assumption is considered on the conditional law, it shifts the focus to the feature distribution $X$ and assumes an extensive knowledge on the distribution of the features. This setting arises naturally in many learning problems. For example, we can get access to distributional information on features in learning scenarios where the sampling mechanism can be controlled, e.g,. in datamodel framework \citep{ilyas2022datamodels}, and gene knockout experiments \citep{peters2016comprehensive,cong2013multiplex}. Other settings include problems where an abundant number of unlabeled data points (unsupervised learning) are available.

The other related line of work is to 
estimate and perform statistical inference on certain statistical model parameters. Specifically, during the past few years, there have been several works \citep{javanmard2014confidence, van2014asymptotically, deshpande2019online, fei2021estimation} for inferential tasks on low-dimensional components of model parameters in high-dimensional $(d>n)$ settings of  linear and generalized linear models. Another complementary line of work, is the conditional independence testing problem $X_j\indep Y| X_{-j}$  to test if a certain feature $X_j$ is independent of the response value $Y$, while controlling for the effect of the other features. This problem has been studied in several recent works for both parametric \citep{crawford2018bayesian,belloni2014inference}, and model-X frameworks \citep{candes2018panning, javanmard2021pearson, liu2022fast, shaer2022learning,berrett2020conditional}. 

Here are couple of points worth mentioning regarding the scope of our paper. 
\begin{enumerate}
    \item \textit{(Feature selection methods)} However Question (*) has a complete different nature from  well-studied variable selection techniques-- with the goal of removing redundant features, an assessment tool provided for (*) can be beneficial for post-processing of feature selection methods as well. Specifically, we expect that two redundant features have close (zero) influence on the response value, therefore our closeness-of-influence test can be used to sift through the set of redundant features and potentially improve the statistical power of the baseline feature selection methods. 
    
    \item \textit{(Regression models)} 
     We would like to emphasize that however fitting any class of regression models would yield an estimate coefficient vector, but comparing the magnitude of coefficient values for answering Question (*) is not statistically accurate and would result in invalid findings, mainly due to model misspecification. Despite such inaccuracies of fitted regression models, our proposed closeness-of-influence test works under no parametric assumption on the conditional law.
    
    \item \textit{(Hardness of non-parametric settings)} The finite-sample guarantee on type-I error rate for our test does not come free. Specifically, this guarantee holds when certain partial knowledge on the feature distributions $\cL(X)$ is known.  
    This setup is often referred as model-X framework \cite{candes2018panning}, where on contrary to the classic statistic setups, the conditional law $\cL(Y|X)$ is optional, and adequate amount of information on features distribution $\cL(X)$ is known. Such requirements for features distribution makes the scope of our work distant from completely non-parametric problems. 
\end{enumerate}

\subsection{Summary of contributions and organization}
In this work, we propose a novel method to test the closeness
of influence of a given pair of features on the response value.
Here is the organization of
the three major parts of the paper:
\begin{itemize}
    \item In Section \ref{section: problem}, we propose the notion of symmetric influence and formulate the question (*) as a tolerance hypothesis testing problem. We then introduce the main algorithm to construct the test statistic, and the decision rule. We later show that the type-I error is controlled for finite number of data points. 
    \item In Section \ref{sec: power}, for two specific learning problems: 1) linear regression setup, and 2) binary classification under a mixture of Gaussians, we analyze the statistical power of our proposed method. Our analysis reveals guidelines on the choice of the score function, that is needed for our procedure. 

    \item In Section \ref{sec: datamodels}, we combine our closeness-of-influence test with datamodels \cite{ilyas2022datamodels} to study the influence of training samples on the trained black box model. We consider CIFAR-10 dataset and identify several pairs of training samples with different influence on the output models. 
    
\end{itemize}
   Finally, we empirically evaluate the performance of our method in several numerical experiments, we show that our method always controls type-I error with finite number of data points, while it can achieve high statistical power. We end the paper by providing concluding remarks and interesting venues for further research. 


\subsection{Notation}
For a random variable $X$, we let $\cL(X)$ denote the probability density function of $X$. For two density functions $p,q$ let $d_\tv(p,q)$ denote the total variation distance. We use $\Phi(t)$ and $\varphi(t)$ respectively for cdf and pdf of standard normal distribution. 
For and integer $n$ let $[n]=\{1,\dots,n\}$ and for a vector $x\in \reals^d$ and integers $i,j\in [d]$ let $x_{\swap(i,j)}$ be a vector obtained by swapping the coordinates $i$ and $j$ of $x$. We let $\normal(\mu,\Sigma)$ denote the probability density function of a multivariate normal distribution with mean $\mu$ and covariance matrix $\Sigma$. 

\section{Problem Formulation}\label{section: problem}
We are interested in investigating that if two given features $i,j$ have close influence on the response value $y$. Specifically, in the case of the linear regression setting $\cL(Y|X)=\normal(X^\sT \th,\sigma^2)$, two features $i$ and $j$ have an equal effect on the response variable $y$, if the model parameter $\th$ has equal coordinates in $i$ and $j$. In this parametric problem, the close influence analysis can be formulated as the following hypothesis testing problem
\[
H_0: |\th_i-\th_j| \leq \tau \,, \quad H_A :|\th_i-\th_j| > \tau\,.
\]
In practice, the considered parametric model may not hold, and due to model misspecification, the reported results are not statistically sound and accurate. Our primary focus is  to extend the definition of close influence of features on the response value to a broader class of supervised learning problems, ideally with no parametric assumption on $\cL(Y|X)$ (model-free). For this end, we first propose the notion of \textit{symmetric influence}.    

\begin{defi}[Symmetric influence]\label{def: influence}
We say that two features $i,j \in [d]$ have a symmetric influence on the response value $y$ if the conditional law $p_{Y|X}$ does not change once features $i$ and $j$ are swapped in $x$. 
More precisely, if $\cL(Y|X)=\cL(Y|X_{\swap(i,j)})$, where $X_{\swap(i,j)}$ is obtained from swapping coordinates $i$ and $j$ in $X$. 
\end{defi}
While the perfect alignment between density function $p_{Y|X}$ and $p_{Y|X_{\swap(i,j)}}$ is considered as equal influence, it is natural to consider small (but nonzero) average distance of these two density functions as having close influence of features $i,j$ on the response value. Inspired by this observation, we cast the problem of closeness-of-influence testing as a tolerance hypothesis testing problem \eqref{eq: null}. Before further analyzing this extended definition, for two simple examples we show that the symmetric influence definition recovers the familiar equal effect notion in parametric problems. It is worth noting that this result can be generalized to a broader class of parametric models.

\begin{propo}\label{propo: glm}
Consider the logistic model $\prob(Y=1|X=x)=\frac{1}{1+\exp(-x^\sT \th)}$. In this model, features $i$ and $j$ have symmetric influence on $y$ if and only if  $\th_i=\th_j$. In addition, for the linear regression setting $y=x^\sT\th+\eps$ with $\eps\sim \normal(0,\sigma^2)$, features $i$ and $j$ have symmetric influence on $y$ if and only if $\th_i=\th_j$. 
\end{propo}
We refer to Appendix A for proofs of all propositions and theorems.

\subsection{Closeness-of-influence testing}
Inspired by the definition of symmetric influence given in Definition \eqref{def: influence}, we formulate the problem of testing the closeness of the influence of two features $i,j$ on $y$ as the following:
\begin{align}\label{eq: null}
\mathcal{H}_0&:~ \E\left[ d_{\tv}(p_{Y|X}, p_{Y|X_{\swap(i,j)}})\right] \leq \tau\,, \nonumber\\
\mathcal{H}_A&:~ \E\left[ d_{\tv}(p_{Y|X}, p_{Y|X_{\swap(i,j)}})\right] > \tau\,. 
\end{align}
Specifically, this hypothesis testing problem allows for general non-negative $\tau$ values. We can test for symmetric influence by simply selecting $\tau=0$. In this case, we must have $p_{Y|X}=p_{Y|X_{\swap(i,j)}}$ almost surely (with respect to some measure on $\cX$). For better understanding of the main quantities in the left-hand-side of \eqref{eq: null}, it is worth to note that $p_{Y|X_{\swap(i,j)}}(y|x)=p_{Y|X}(y|x_\swap(i,j))$ and the quantity of interest can be written as
\begin{align*}
&\E\left[ d_{\tv}(p_{Y|X}, p_{Y|X_{\swap(i,j)}})\right]\\
&=\frac{1}{2}\int\Big|p_{Y|X}(y|x)-p_{Y|X}(y|x_{\swap(i,j)})\Big|p_X(x)\de y \de x\,. \end{align*}
We next move to the formal process to construct the test statistics of this hypothesis testing problem.

\noindent\textbf{Test statistics}. We first provide high-level intuition behind the test statistics used for testing \eqref{eq: null}. In a nutshell, for two i.i.d. data points $(x^{(1)},y^{(1)})$ and $(x^{(2)},y^{(2)})$, if the density functions $p_{Y|X}$ is close to $p_{Y|X_{\swap(i,j)}}$, then for an optional score functions applied on $(x^{(1)},y^{(1)})$ and $(x_{\swap(i,j)}^{(2)},y^{(2)})$, with equal chance (50$\%$) one should be larger than the other one. This observation is subtle though. Since we intervene in the features of the second data point (by swapping its coordinates), this shifts the features distribution, thereby the joint distribution of $(x^{(1)},y^{(1)})$ and $(x_{\swap(i,j)}^{(2)},y^{(2)})$ are not equal. This implies that we must control for such distributional shifts on features as well. The formal process for constructing the test statistics $U_n$ is given in Algorithm~\ref{algorithm: importance}. We next present the decision rule for hypothesis problem \eqref{eq: null}.

\begin{algorithmic}[t]
	\begin{algorithm}
			\textbf{Input:}
			$n$ data points $\{(x^{(m)},y^{(m)})\}_{m=1:n}$ with $(x,y) \in \reals^d \times \reals$ (for $n$ being even--if not, remove one sample), two features  $i,j \in \{1,2,...,d\}$, and a score function $T:\mathcal{X}\times \mathcal{Y}\rightarrow \reals$.
			
		\smallskip
		\textbf{Output:} 
		A test statistic $U_{n}$\,. \\
\medskip

 \STATE For $1\leq m\leq \frac{n}{2}$ define
\[
\tx^{(m)}=x^{(m+\frac{n}{2})}_{\swap(i,j)}\,, \quad \ty^{(m)}=y^{(m+\frac{n}{2})}\,.
\]
\STATE Define tests statistic $U_{n}$
\[
U_n=\frac{2}{n}\sum_{m=1:\frac{n}{2}} \ind\left(T\big(x^{(m)},y^{(m)}\big)\geq T\big(\tx^{(m)},\ty^{(m)}\big)\right)\,.
\]

\caption{Test statistic for hypothesis testing \eqref{eq: null}}\label{algorithm: importance}
	\end{algorithm}
\end{algorithmic}

\noindent\textbf{Decision rule}. For the data set $(\bX,\bY)$ of size $n$ and test statistic $U_n$ as per Algorithm \ref{algorithm: importance} at significance level $\alpha$ consider the following decision rule
\begin{equation}\label{eq: decision}
\psi_n (\bX,\bY) =\ind\left(\Big|U_n-\frac{1}{2}\Big|\geq \tau+ \tau_{X}+\sqrt{\frac{\log (2/\alpha)}{n}}\right)\,,
\end{equation}
with $\tau_X$ being an upper bound on the total variation distance between the original feature distribution, and the obtained distribution by swapping coordinates $i,j$. More precisely, for two independent features vectors $X^{(1)},X^{(2)}$ let $\tswap$ be such that $\tswap\geq d_{\tv}\left(\cL(X^{(1)}) ,\cL(X^{(2)}_{\swap(i,j)})  \right)$. In fact, in several learning problems when features have a certain symmetric structure, the quantity $\tau_X$ is zero. For instance, when features are multivariate Gaussian with isotropic covariance matrix. More on this can be seen in Section \ref{sec: swap}. 

\noindent\textbf{Size of the test}. 
In this section, we show that the obtained decision rule \ref{eq: decision} has control on type I error with finite number of samples. More precisely, we show that the probability of falsely rejecting the null hypothesis \eqref{eq: null} can always be controlled such that it does not exceed a predetermined significance level $\alpha$. 
\begin{thm}\label{thm: size}
Under the null hypothesis \eqref{eq: null}, decision rule \eqref{eq: decision} has type-I error smaller than $\alpha$. More precisely
\[
\prob_{\mathcal{H}_0}(\psi(\bX,\bY)=1)\leq \alpha\,.
\]
\end{thm}
Based on decision rule \eqref{eq: null}, we can construct p-values for the hypothesis testing problem \eqref{eq: null}. The next proposition gives such formulation. 
\begin{proposition}\label{propo: p-val}
Consider
\begin{equation}\label{eq: p-val}
p=\begin{cases}
1\,, & |U_n-1/2|\leq \tau+\tau_X\,,\\
1\wedge \eta_n(U_n,\tau,\tau_{X}) \,,& \text{otherwise} \,,
\end{cases}
\end{equation}
with function $\eta_n(u,\tau_1,\tau_2)$ being defined as
\[
\eta_n(u,\tau_1,\tau_2)=2\exp\left(-n\bigg(\Big|u-\frac{1}{2}\Big|-\tau_1-\tau_2\bigg)^2\right)\,.
\]
 In this case, the p-value $p$ is super-uniform. More precisely, under the null hypothesis \eqref{eq: null} for every $\alpha\in [0,1]$ we have
\[
\prob(p\leq \alpha)\leq \alpha\,.
\]
\end{proposition}
\subsection{Effect of feature swap on features distribution}\label{sec: swap}
From the formulation of the decision rule given in \eqref{eq: decision}, it can be seen that an upper bound on total variation distance between density functions of $X^{(1)}$ and $X^{(2)}_{\swap(i,j)}$ is required. This quantity shows up as $\tau_X$ in \eqref{eq: decision}. Regarding this change on $X$ distribution, two points are worth mentioning. First, in several classes of learning problems the feature vectors follow a symmetric structure which renders the quantity $\tau_X$ to zero. For instance, when features have an isotropic Gaussian distribution (Proposition \ref{propo: gaussian-swap}), or in the datamodel sampling scheme \citep{ilyas2022datamodels}, the formal statement is given in Proposition \ref{propo: unif-binary}. 
Secondly, the value of $\tau_X$ can be computed when adequate amount of information is available on distribution of $X$, the so-called model-X framework \citep{candes2018panning}. 
We would also like to emphasize that indeed we do not need the direct access to entire density function $p_X$ information,  and an upper bound on the quantity $d_\tv(\cL(X^{(1)}), \cL(X^{(2)}_{\swap(i,j)}))$ is sufficient. In the next proposition, for the case that features follow a general multivariate Gaussian distribution $\normal(\mu,\Sigma)$ we provide a valid closed-form value for $\tau_X$. 

\begin{propo}\label{propo: gaussian-swap}
Consider a multivariate Gaussian distribution with the mean vector $\mu\in \reals^d$ and the covariance matrix $\Sigma\in \reals^{d\times d}$, for two features $i$ and $j$ the following holds:
\begin{multline}
d_{\tv}\left(\cL(X^{(1)}),\cL(X^{(2)}_{\mathsf{swap}(i,j)})\right)\\ \leq
\frac{1}{2}\Big[\trace\big(-I_d+P_{ij}\Sigma^{-1}P_{ij}\Sigma\big) \\ 
+(\mu-P_{ij}\mu)^\sT \Sigma^{-1}(\mu-P_{ij}\mu) \Big]^{1/2}\,,
\end{multline}
where $P_{ij}$ is the permutation matrix that swaps the coordinates $i$ and $j$. More precisely, for every $x\in \reals^d$
we have $P_{ij}x=x_{\swap(i,j)}$. 
\end{propo}
It is easy to observe that in the case of isotropic Gaussian distribution with zero mean, we can choose $\tau_X=0$. More concretely, when $\mu=0$, and $\Sigma=\sigma^2I$, then Proposition \ref{propo: gaussian-swap} reads $\tau_X=0$. 
We next consider a setting with binary feature vectors that arise naturally in datamodels \citep{ilyas2022datamodels}, and will be used later in experiments of Section \ref{sec: datamodels}. 
\begin{propo}\label{propo: unif-binary}
Consider a learning problem with binary features vector $x\in \{0,1\}^d$. For a positive integer $m$, we suppose that $x$ is sampled uniformly at random from the space $S_m=\{x\in \{0,1\}^d: \sum x_i=m\}$. This means that the output sample has binary entries with exactly $m$ non-zero coordinates. Then, in this setting for two independent features vectors $x^{(1)},x^{(2)}$, the following holds
\[
d_{\tv}\left(\cL\big(X^{(1)}\big),\cL\big(X^{(2)}_{\mathsf{swap}(i,j)}\big)\right)=0\,.
\]
\end{propo}
\section{Power Analysis}\label{sec: power}
In this section, we provide a power analysis for our method. 
For a fixed score function $T:\mathcal{X}\times \mathcal{Y}\rightarrow \reals$ and two i.i.d. data points $(x^{(1)},y^{(2)})$ and $(x^{(2)},y^{(2)})$ consider the following cumulative distribution functions:
\begin{align*}
F_T(t)&=\prob\left(T(X^{(1)},Y^{(1)})\leq t\right)\,,\\ 
G_T(t)&=\prob\left(T(X^{(2)}_{\swap(i,j)},Y^{(2)})\leq t\right)\,.
\end{align*}
In the next theorem, we show that the power of our test depends on the average deviation of the function  $F_T\circ G_T^{-1}$ from the identity mappinp on the interval $[0,1]$. 
\begin{thm}\label{thm: power}
Consider the hypothesis testing problem \eqref{eq: null} at significance level $\alpha$ with $n$ data points $(\bX,\bY)$. In addition, suppose that score function $T:\mathcal{X}\times \mathcal{Y}\to \reals$ satisfies the following condition  for some $\beta \in (0,1)$:
\[
\left| \int_0^1 (F_T(G_T^{-1}(u))-u) \de u \right | \geq \rho_{n}(\alpha,\beta,\tau)+\tau_{X} \,,
\]
with $\rho_n(\alpha,\beta,\tau)=2\exp(-n\beta^2)+\sqrt{\frac{\log(2/\alpha)}{n}}+\tau$.
In this case, the decision rule \eqref{eq: decision} used with the score function  $T$ has type II error not exceeding $\beta$. More precisely $
\prob\left(\Psi_n(\bX,\bY)=1\right) \geq 1-\beta\,.
$
\end{thm}
The function $F_T\circ G_T^{-1}$ is called  \textit{ordinal dominance curve} (ODC) \cite{hsieh1996nonparametric,bamber1975area}. It can be seen that the ODC is the population counterpart of the PP plot.  A direct consequence of the above theorem is that if the ODC has a larger distance from the identity map $i(u)=u$, then it would be easier for our test to flag smaller gaps between the influence of features. We next focus on two learning problems: 1) linear regression setting, and 2) binary classification under Gaussian mixture models. For each problem, we use Theorem \ref{thm: power} and provide lower bounds on the statistical power of our closeness-of-influence test.
\newline\noindent\textbf{Linear regression setup}.
In this setting, we suppose that $y=x^\sT \th^*+\eps$ for $\eps\sim\normal(0,\sigma^2)$ and feature vectors drawn iid from a multivariate normal distribution $\normal(0,I_d)$. Since features are isotropic Gaussian with zero mean, by an application of Theorem \ref{propo: gaussian-swap} we know that $\tau_X$ is zero. In the next theorem, we provide an upper bound for hypothesis testing problem \eqref{eq: null} with $n$ data points and the score function $T(x,y)=|y-x^\sT\hth|$ for some model estimate $\hth$. We show that in this example, the power of the test highly depends on the value $|\th^*_i-\th^*_j|$ and the quality of the model estimate $\hth$. Indeed, the higher the contrast between the coefficient values $\th^*_i$ and $\th^*_j$, the easier it is for our test to reject the null hypothesis.   
\begin{thm}\label{thm: power-linear}
Under the linear regression setting $y=x^\sT \th^*+\eps$ with $\eps\sim\normal(0,\sigma^2)$ with feature vectors coming from a normal population $x\sim \normal(0,I_d)$, consider  the hypothesis testing problem \eqref{eq: null} for features $i$ and $j$ with $\tau\in (0,1)$. We run Algorithm \ref{algorithm: importance} at the significance level $\alpha$ with the score function $T(x,y)=|y-x^\sT\hth|$ for a model estimate $\hth \in \reals^d$. For $\beta\in (0,1)$ such that $\tan(\frac{\pi}{2}\rho_n(\alpha,\beta,\tau))\leq \frac{1}{2}$, suppose that the following condition holds
{\small\[
|\th^*_i-\th^*_j|\geq \frac{2\tan(\frac{\pi}{2}(\rho_n(\alpha,\beta,\tau)))}{1-2\tan(\frac{\pi}{2}(\rho_n(\alpha,\beta,\tau))) }\frac{\Big(\sigma^2+\|\hth-\th^*\|_2^2 \Big)}{|\hth_i-\hth_j|}\,,
\]}
for $\rho_n(\alpha,\beta,\tau)$ as per Theorem \ref{thm: power}.
Then, the type II error is bounded by $\beta$. More precisely, we have $
\prob(\Psi_n(\bX,\bY)=1)\geq 1-\beta\,.$
\end{thm}

We refer to Appendix for the proof of Theorem \ref{thm: power-linear}. It can be seen that the right-hand-side of the above expression can be decomposed into two major parts. The first part involves the problem parameters, such as the number of samples $n$, and error tolerance values $\alpha$ and $\beta$. This quantity for a moderately large number of samples $n$, and small tolerance value $\tau$ can get sufficiently small.  On the other hand, the magnitude of the second part depends highly on the quality of the model estimate $\hth$ and the inherent noise value of the problem $\sigma^2$ which basically indicates how structured is the learning problem. Another interesting observation is regarding the $|\hth_i-\hth_j|$. Indeed, it can be inferred that small values of this quantity renders the problem of discovering deviation from the symmetric influence harder. This conforms to our expectation, given that in the extreme scenario that $\hth_i=\hth_j$ it is impossible for the score function to discern $\th^*_i$ and $\th^*_j$, because of the additive nature of the considered score function.

\noindent\textbf{Binary classificaiton}. In this section,  we provide power analysis of our method for a binary classification setting. Specifically, we consider the binary classification under a mixture of Gaussian model. More precisely, in this case the data generating process is given by
\begin{equation}\label{eq: GMM}
y=\begin{cases} +1\,,&\text{w.p } q\,,\\
 -1 \,, & \text{w.p } 1-q\,. \end{cases}\,, \quad x\sim \normal(y\mu,I_d)\,.
\end{equation}
We consider the influence testing problem \eqref{eq: null} with $\tau=0$.
In the next theorem, we provide a lower bound on the statistical power of our method used under this learning setup.  

\begin{thm}\label{thm: power-binary}
Under the binary classification setup \eqref{eq: GMM}, consider the hypothesis testing problem \ref{eq: null} for $\tau=0$. We run Algorithm \ref{algorithm: importance} with the score function $T(x,y)=yx^\sT \th$ at the significance level $\alpha$, and suppose that for some nonnegative value $\beta$ the following holds
\[
|\mu_i-\mu_j|\geq \Phi^{-1}\left(\frac{1}{2}+\rho_n(\alpha,\beta,0) \right)\frac{\sqrt{2}\|\hth\|_2}{|\hth_i-\hth_j|}\,,
\]
where $\rho_n(\alpha,\beta,\tau)$ is given as per Theorem \ref{thm: power}. Then the type-II error in this case is bounded by $\beta$. More concretely, we have $
\prob(\Psi_n(\bX,\bY)=1)\geq 1-\beta\,.$

\end{thm}


It is important to note that in this particular setting, the features do not follow a Gaussian distribution with a zero mean. Instead, they are sampled from a mixture of Gaussian distributions with means $\mu$ and $-\mu$. The reason why $\tau_X=0$ can be utilized is not immediately obvious. However, we demonstrate that when testing for $\tau=0$ under the null hypothesis, it is necessary for $\mu_i$ to be equal to $\mu_j$, and the distribution of features remains unchanged when the coordinates $i$ and $j$ are swapped. As a result, we can employ $\tau_X=0$ in this scenario. This argument is further elaborated upon in the proof of Theorem \ref{thm: power-binary}.

From the above expression it can be observed that for sufficiently large number of data points $n$ and a small value $\tau$, the value $\Phi^{-1}(1/2+\rho_n)$ will get smaller and converge to zero. In addition, it can be inferred that an ideal model estimate $\hth$ must have small norm and high contrast between $\hth_i$ and $\hth_j$ values. An interesting observation can be seen on the role of other coordinate values in $\hth$. In fact, it can be realized that for the choice of the score function $T(x,y)=yx^\sT \hth$, the support of the model estimate $\hth$ must be a subset of two features $i$ and $j$, since this would decrease $|\hth|$ and increases the value of $|\hth_i-\hth_j|$.

\section{Experiments} \label{sec: exp}

\begin{figure*}[t]
\centering
\subfigure[Average rejection rate of the null hypothesis of \eqref{eq: null} for $\tau=0$ and  features with isotropic Gaussian distribution $x\sim \normal(0,I_{10})$. In this experiment, we consider $y|x\sim\normal(x^\sT \th^*,1)$ for $\th^*= (1,1,2,2,3,3,4,4,5,5)$. 
]{ 
\includegraphics[width=0.45\linewidth]{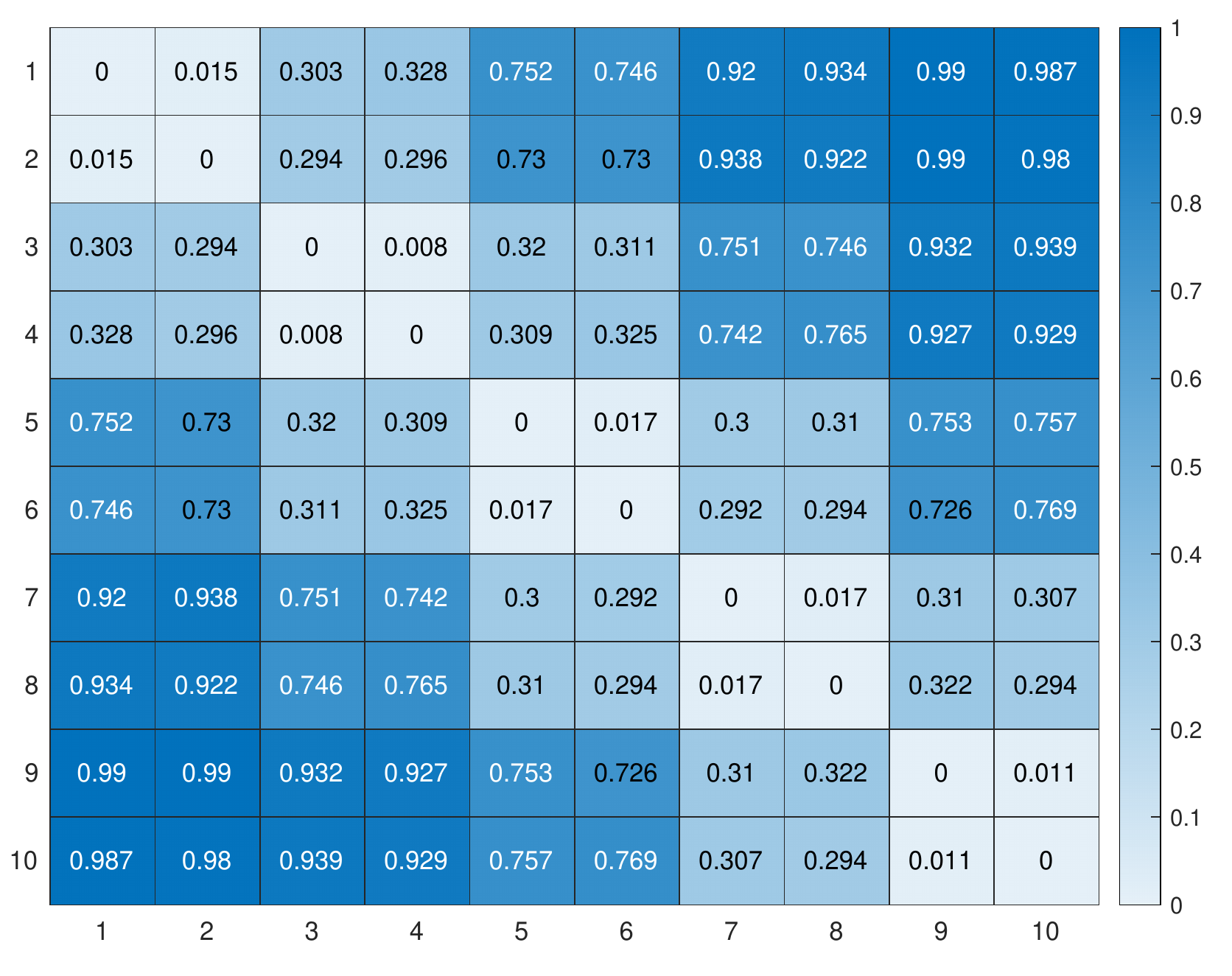}
\label{fig: power-isotropic}
}
\hfill
\subfigure[Average rejection rate of the null hypothesis \eqref{eq: null} for $\tau=0$ and features coming from an isotropic Gaussian distribution $x\sim \normal(0,I_{10})$. In this experiment, we consider $y|x \sim \normal(x^\sT S x,1)$ for a positive definite matrix $S_{i,j}=1+\ind(i=j)$ ($2$ on diagonal and $1$ on off-diagonal entries). 
]{
\includegraphics[width=0.45\linewidth]{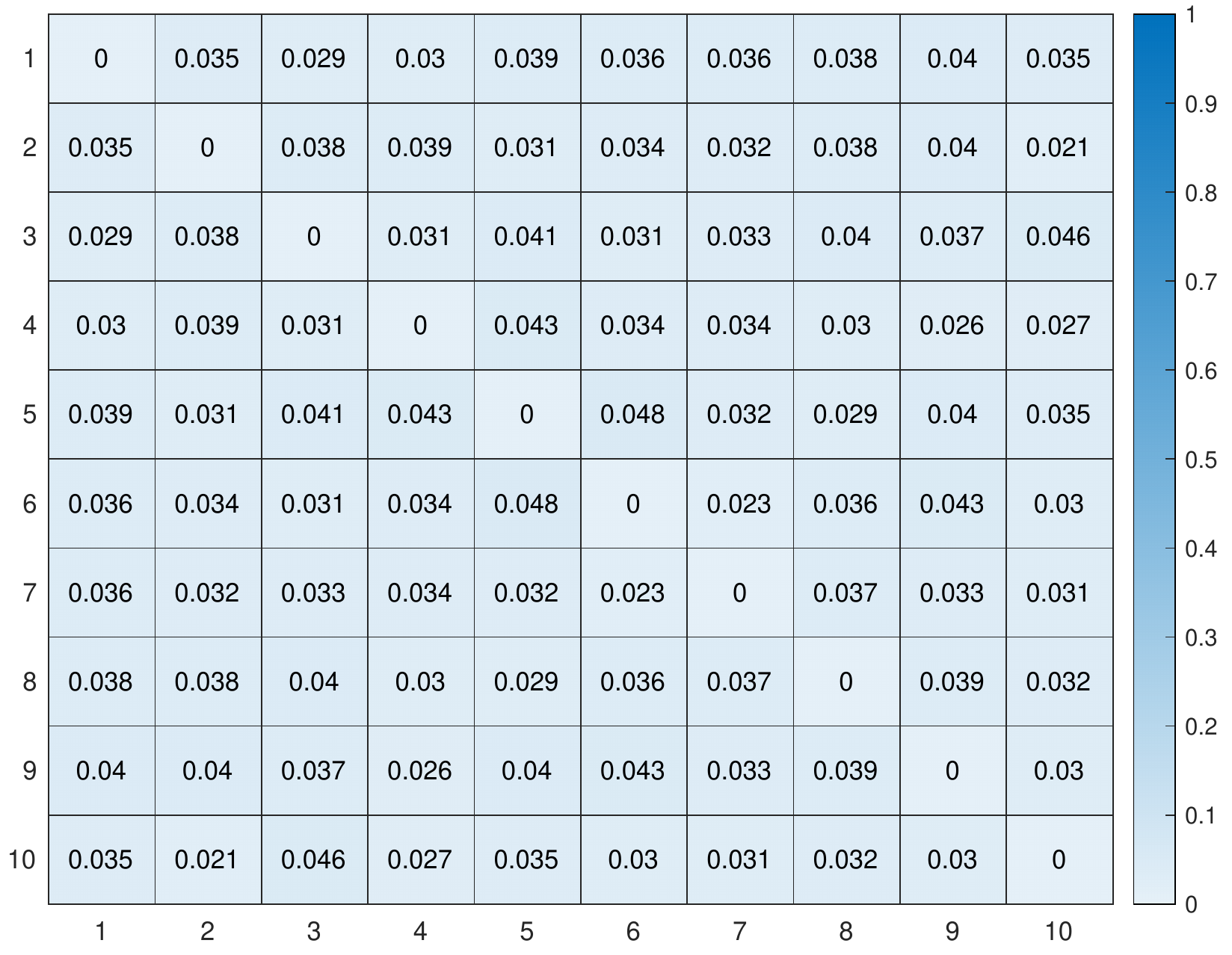}
\label{fig: size-isotropic}
}
\caption{Average Rejection Rates for Different Settings}
\hspace{-7mm}

\vspace{-2mm}
\end{figure*}

In this section, we evaluate the performance of our proposed method for identifying the symmetric influence across features. We start by the Isotropic Gaussian model for feature vectors. More precisely, we consider $x\sim \normal(0,I_d)$ with $d=10$. In this case, we have $\tau_X=0$ and we consider the hypothesis testing problem \eqref{eq: null} for $\tau=0$ (symmetric influence). 
\newline\noindent\textbf{Size of the test.} We first start by examining the size of our proposed method. For this end, we consider the conditional law $y|x\sim \normal(x^\sT S x,1)$, for a semi-positive definite matrix $S$ with coordinate $(i,j)$ being $S_{i,j}=1+\ind(i=j)$. The conditional mean of $y|x$ is a quadratic form and it is easy to observe that in this case for every two features $i,j\in \{1,\dots,10\}$ we have 
$x^\sT S x=x_{\swap(i,j)}^\sT S x_{\swap(i,j)}$, and therefore the null hypothesis holds.  We test for the symmetric influence of each pair of features ($\binom{10}{2}$ number of tests). We run our method with the score function $T(x,y)=|y-\hth^\sT x|$ with $\hth\sim \normal(0,I_d)$. The estimate $\hth$ is fixed across all $45$ tests. We suppose that we have access to $1000$ data points, and we consider three different significance levels $\alpha=0.1,0.15,$ and $0.2$. The results of this experiment can be seen in Figure \ref{fig: size-isotropic} where the reported numbers (rejection rates) are averaged over $1000$ independent experiments. It can be observed that, in this case for all three significance levels, the rejection rates are smaller than $\alpha$, and therefore the size of the test is controlled.

\noindent\textbf{Power analysis.} The linear regression setting is considered, in which $y|x\sim \normal(x^\sT \th^*, 1)$, for $\th^*\in \reals^d$ with $d=10$. We consider the following pattern for signal strength $\th^*_1=\th^*_2=1$, $\th^*_3=\th^*_4=2$, $\th^*_5=\th^*_6=3$, $\th^*_7=\th^*_8=4$, $\th^*_9=\th^*_{10}=5$. In this example, it can be observed that the following pairs of features $\cI=\{(1,2),(3,4),(5,6),(7,8),(9,10)\}$ have symmetric influence, and for any other pair the null hypothesis \eqref{eq: null} must be rejected. We use the score function $T(x,y)=|y-x^\sT \hth|$ at significance level $\alpha=0.1$ for three different choices of $\hth$. We follow this probability distribution $\hth\sim\normal(\th_0,\sigma^2 I_d)$ for three different $\sigma$ values $\sigma=1,2,$ and $3$. A smaller value of $\sigma$ implies a better estimation of $\th_0$. The average rejection rates are depicted in Figure \ref{fig: power-isotropic}, where each $10\times 10$ square corresponds to a different $\sigma$ value (three plots in total). Specifically, $(i,j)$-th cell in each plot denotes the average rejection rate of the symmetric influence hypothesis for features $i$ and $j$. The rejection rates are obtained by averaging over $1000$ independent experiments. First, it can be inferred that for pairs belonging to the set $\cI$ the rejection rate is always smaller than the significance level $\alpha=0.1$, thereby the size of the test is controlled. In addition, by decreasing the $\sigma$ value (moving from right to left), it can be inferred that the test achieves higher power (more dark blue regions). It is consistent with our prior expectation that the statistical power of our method depends on the quality of the score function $T$ and model estimate $\hth$; see Theorem \ref{thm: power-linear}. More on the statistical power of our method, it can be observed that within each plot, pairs that have higher contrast in the difference of coefficient magnitudes have higher statistical power. For instance, this pair of features $(1,10)$ with coefficient values $\th^*_1=1,\th^*_{10}=5$ has rejection rates of $0.987,0.768 ,0.543$ (for $\sigma=1,2,3$, respectively) while the other pair of features $(6,8)$ with coefficient values $\th^*_6=3,\th^*_{8}=4$ has rejection rate of $0.294,0.097,0.055$ (for $\sigma=1,2,3$, respectively).

\begin{figure}[t]
\centering
\includegraphics[width=0.7\linewidth]{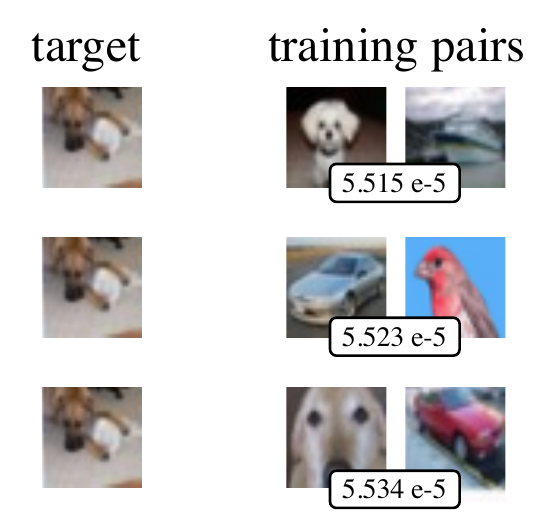}
\caption{Summary of discoveries on CIFAR-10 dataset via datamodels used with our closeness-of-influence test. For each pair of 10 classes (can be similar), we choose random samples from the training data along with a random target image from dog pictures in the test data, and we repeat this process 20 times. After running the  Benjamini–Yekutieli procedure on output p-values (2000 in total) at $\alpha=0.2$ three significant results are reported. The images of these findings are plotted above, with their associated p-values. This implies that with high certainty images in each pair influence the target example differently. }
\label{fig: findings}
\end{figure}

\section{Influence of Training Data on Output Model}\label{sec: datamodels}

In this section, we combine our closeness-of-influence test with datamodel framework \cite{ilyas2022datamodels} to analyze the influence of training samples on the evaluations of the trained model on certain target examples. We first provide a brief overview on datamodels and later describe the experiments setup.

\subsection{Datamodels}
For training samples $\cD^{\mathsf{train}}=\{(x_i,y_i)\}_{i=1:N}$ consider a class of learning algorithm $\cA$, where by class we mean a training mechanism (potentially randomized), such as training a fixed geometry of deep neural networks via gradient descent and a fixed random initialization scheme. In datamodels \cite{ilyas2022datamodels}, a new learning problem is considered, where feature vectors $S$  are binary 0-1 vectors with size $N$ with $\gamma\in (0,1)$ portion one entries, selected uniformly at random. Here $S$ is an indicator vector for participation of $N$ data points $\cD^{\mathsf{train}}$ in the training mechanism, i.e,. $S_i=1$ if and only if the $i-$th sample of $\cD^{\mathsf{train}}$ is considered for the training purpose via $\cA$. For a fixed target example $x$, the response value is the evaluation (will be described later) of the output model (trained with samples indicated in $S$) on $x$, denoted by $f_{\cA}(x;S)$. This random sampling of data points from $\cD^{\mathsf{train}}$ is repeated $m$ times, therefore data for the new learning problem is $\{(S_i,f_{\cA}(x,S_i))\}_{i=1:m}$. The ultimate goal of datamodels is to learn the mapping $S\to f_{\cA}(x,S)$ via surrogate modeling and a class of much less complex models. In the seminal work of \cite{ilyas2022datamodels}, they show that using linear regression with $\ell_1$ penalty (LASSO \cite{tibshirani1996regression}) performs surprisingly well in learning the highly complex mapping of $S\to f_{\cA}(x,S)$.
\subsection{Motivation}
We are specifically interested in analyzing the influence of different pairs of training samples on a variety of test targets, and discover pairs of training samples that with high certainty influence the test target differently. We use the score function $(f_{\cA}(x,S)-x^\sT \hth)^2$ for our closeness-of-influence test, where $\hth$ is the learned datamodel. We adopt this score function, mainly due to the promising performance of linear surrogate models in \cite{ilyas2022datamodels} for capturing the dependency rule between $S$ and $f_{\cA}(x;S)$. In addition, the described sampling scheme in datamodels  satisfies the symmetric structure as per Proposition \ref{propo: unif-binary} (so $\tau_X=0$). We would like to emphasize that despite the empirical success of datamodels, the interpretation of training samples with different coefficient magnitude in the obtained linear datamodel $\hth$ is \textit{not} statistically accurate. Here we approach this problem through the lens of hypothesis testing and output p-values, to project the level of confidence in our findings. 
\subsection{Experimental Setups and Results}
We consider the CIFAR-10 dataset \cite{krizhevsky2009learning}, which has $N=50000$ training samples along with $10000$ test datapoints and 10 classes  \footnote{{airplane, automobile, bird, cat, deer, dog, frog, horse, ship, and truck}}. We consider $\gamma=0.5$ (portion of ones in $S_i$ samples), and follow the same heuristics provided for $f_\cA(x;S)$ in  \cite{ilyas2022datamodels}, which is the correct-class margin, defined as the logit value of the true class minus the highest logit value among incorrect classes. We use the datamodel data given in \url{https://github.com/MadryLab/datamodels-data}. The provided data has $310k$ samplings, where for each target example $x$ (in the test data) the datamodel parameter $\hth\in \reals^N$ is estimated via the first $300k$ samples ($10000$ total number of datamodels $\hth$ for each test data).  We use the additional $10k$ samples to run our closeness-of-fit test with the linear score function $(f_{\cA}(x;S)-x^\sT \hth)^2$.  Now, for each pair of training samples and a specific target test example, we can test for their closeness of influence. In the first experiment, for each two classes (can be the same) we choose two pictures as the training pair (randomly from the two classes), and for the target sample, we select randomly from the class of dog pictures. For each two classes, we repeat this process $20$ times, and run our test \eqref{eq: null} with $\tau=0$, and report all p-values ($2000$ in total). After running the Benjamini–Yekutieli procedure \cite{benjamini2001control} (with log
factor correction to control for dependency among p-values), we find three statistically significant results at $\alpha=0.2$ with p-value=$5\times 10^{-5}$ (for all three discoveries).  Surprisingly, all three findings correspond to a similar test image, the pictures of training pairs and the one test image can be seen in Figure~\ref{fig: findings}.
It can be observed that in all findings one of the reported images is visually closer to the target image. This conforms well to obtained results that the null hypothesis \eqref{eq: null} which states that the two  training images have equal influence on the target sample is rejected. We refer to  Appendix B for the rest of experiments.



\section{Concluding Remarks}
In this paper, we proposed a novel method to test the closeness of influence of a given pair of features on the response value. This procedure makes no assumption on the conditional law between the response value and features ($\mathcal{L}(Y|X)$). We first proposed a notion called "symmetric influence" that generalized the familiar concept of equal coefficient in parametric models. This notion is motivated to characterize the sensitivity of the conditional law with respect to swapping the features. We then formulated the closeness-of-influence testing problem as a tolerance hypothesis testing. We provide theoretical guarantees on type-I error rate. 
We then analyzed statistical power of our method for a general score function $T$, and show that for two specific learning problems i) linear regression settings, and 2) binary classification under a mixture of Gaussian models with a certain choice of score functions we can achieve full statistical power. Finally, we adopt the datamodel framework and use our closeness-of-influence test to find training samples that have different influence on the trained model.

Several interesting venues for future research are in order. In particular, extending this framework for multiple testing (testing for multiple number of pairs) and still achieving valid statistical results. This can be done with generic multiple testing frameworks (similar to Benjamini–Yekutieli procedure used in Section \ref{sec: datamodels}) on the obtained p-values, but a method that is crafted for this setting can be more powerful. In addition, extending this framework for studying influence of a group of features (more that two) can be of great interest.

\balance
\bibliographystyle{icml2023}
\bibliography{mybib}

\input{supplement2}



\end{document}

%% file: math_commands.tex
\usepackage{amsmath,amsfonts,bm}

\def\eqref#1{\ref{#1}}

\def\1{\bm{1}}

\def\eps{{\epsilon}}

\DeclareMathAlphabet{\mathsfit}{\encodingdefault}{\sfdefault}{m}{sl}
\SetMathAlphabet{\mathsfit}{bold}{\encodingdefault}{\sfdefault}{bx}{n}

\def\tX{{\tens{X}}}

\def\sT{{\mathbb{T}}}

\newcommand{\E}{\mathbb{E}}



%% file: preamble.tex
\usepackage{xcolor}
\definecolor{thedarkblue}{RGB}{0,0,120} 
\definecolor{mydarkblue}{rgb}{0,0.08,0.45} 
\definecolor{darkblue}{rgb}{0,0.08,180}
\colorlet{TufteRed}{red!80!black}
\definecolor{theblue}{RGB}{0,0,180}
\colorlet{thered}{TufteRed}
      
\usepackage{microtype}
\usepackage{balance}
\usepackage{booktabs}
\usepackage{tabularx}

\usepackage{amsmath,amssymb,amsthm}

\newcommand{\eat}[1]{\ignorespaces}
\usepackage{comment}

\usepackage{tikz}
\usepackage{verbatim}
\usetikzlibrary{arrows}
\usetikzlibrary{shapes,snakes}
\usetikzlibrary{decorations.pathmorphing} 
\usetikzlibrary{fit}					
\usetikzlibrary{backgrounds}	

\usepackage{ragged2e}
\usepackage{multirow}
\usepackage{microtype}
\usepackage{balance}
\usepackage{setspace}

\graphicspath{{./}{./graphics/}}
\newcolumntype{H}{>{\setbox0=\hbox\bgroup}c<{\egroup}@{}}

\newcolumntype{R}[1]{>{\RaggedLeft\arraybackslash}} 
\newcolumntype{L}[1]{>{\RaggedRight\arraybackslash}} 

\newtheorem{theorem}{Theorem}[section]

\newtheorem{lemma}[theorem]{Lemma}

\providecommand{\eps}{\varepsilon}%

\DeclareMathOperator{\hugeE}{\mbox{\huge\raise-0.3ex\hbox{E}}}
\DeclareMathOperator{\p}{\mathbb{P}}
\DeclareMathOperator{\hugep}{\mbox{\huge\raise-0.3ex\hbox{$\p$}}}

\providecommand{\prob}[1][]{\p_{#1}\BraceOf}

\DeclareMathOperator{\trace}{trace}

\providecommand{\eps}{\varepsilon}

\providecommand{\tX}{\ensuremath{\tensor{X}}}

\usepackage{subfigure}
\usepackage{graphbox}
\usepackage{svg}
\usepackage{nicefrac}

\DeclareMathAlphabet{\mathbcal}{OMS}{cmsy}{b}{n}
\usepackage{amsmath}
\usepackage{mathrsfs}
\usepackage{comment}
\usepackage{bm}
\usepackage{bbm}
\usepackage{amssymb}
\usepackage{enumitem}

%% file: notation.tex
\def\tX{\widetilde{X}}

\def\tX{\widetilde{X}}
\def\tx{\widetilde{x}}

\def\cA{{\cal A}}

\def\cD{{\cal D}}

\def\reals{{\mathbb R}}

\def\eps{{\varepsilon}}
\def\prob{{\mathbb P}}
\def\E{{\mathbb E}}

\def\L0{{L_i}}

\def\de{{\rm d}}
\def\<{\langle}
\def\>{\rangle}

\def\hth{\widehat{\theta}}

\def\F{{\sf F}}
\def\ind{{\mathbb I}}

\def\F{{\sf F}}
\def\normal{{\sf N}}

\def\tX{{\widetilde{X}}}

\def\sT{{\sf T}}

\def\v*{v_i}
\def\T*{T_i}

\def\u*{u_i}
\def\F*{F_i}

\def\cI{{\mathcal{I}}}

\def\ty{\tilde{y}}
\def\tx{\tilde{x}}

\def\cL{\mathcal{L}}

\def\th{\theta}
\def\hth{{\widehat{\theta}}}

\def\cX{\mathcal{X}}

\def\tmu{\widetilde{\mu}}

\def\l1u{W}

\newcommand{\ajcomment}[1]{}

\makeatletter
\newcommand{\labitem}[2]{%
\def\@itemlabel{\text{#1}}
\item
\def\@currentlabel{#1}\label{#2}}
\makeatother

\DeclareMathAlphabet{\mathpzc}{OT1}{pzc}{m}{it}


%% file: supplement2.tex
\newpage
\appendix
\onecolumn

\newcommand{\tsigma}{\widetilde{\sigma}}

\section{Proof of Theorems and Technical Lemmas}

\subsection{Proof of Theorem \ref{thm: size}}\label{sec: proof-size}

Consider two data points $z^{(1)}=(x^{(1)},y^{(1)})$, $z^{(2)}=(x^{(2)},y^{(2)})$ drawn i.i.d. from the density function $p_{X,Y}$. For two features $i,j$, define
\[
\pi=\prob\left(T\big(X^{(1)},Y^{(1)}\big) \geq  T\big(X^{(2)}_{\swap(i,j)},Y^{(2)}\big) \right)\,.
\]
We want to show that under the null hypothesis, the value $\pi$ is concentrated around $1/2$ with maximum distance of $\tau_X$.  
First, from the symmetry between two i.i.d. data points we have
\begin{align*}
 \prob\left(T\big(X^{(1)},Y^{(1)}\big)\geq T\big(X^{(2)},Y^{(2)}\big)\right)=1/2\,.
\end{align*}
The underlying assumption is that in the case of equal values the tie is broken randomly. We introduce  $\widetilde{z}^{(2)}=(x^{(2)}_{\swap(i,j)},y^{(2)})$. This brings us
\small{
\begin{align*}
   \pi-\frac{1}{2}&=\prob\left(T\big(X^{(2)}_{\swap(i,j)},Y^{(2)}\big)\leq T(Z_1)\right)\\
   &\,-\prob\left(T\big(X^{(2)},Y^{(2)}\big)\leq T(Z_1)\right)\\
   &=\E\left[\prob(T(\widetilde{Z}^{(2)})\leq T(Z^{(1)})|Z^{(1)},Y^{(2)})\right]\\
   &\,-\E\left[\prob(T(Z^{(2)})\leq T(Z^{(1)})|Z^{(1)},Y^{(2)})\right]\,.
\end{align*}
}
In the next step, we let $T^{(1)}=T(Z^{(1)})$, $T^{(2)}=T(Z^{(2)})$, and $\widetilde{T}^{(2)}=T(\tz^{(2)})$. Then, by an application of Jenson's inequality we get
\begin{align}
\Big|\pi-\frac{1}{2}\Big|    &\leq \E\Big[\big|\prob(\widetilde{T}^{(2)}\leq T^{(1)}|Z^{(1)},Y^{(2)})-\prob({T}^{(2)}\leq T^{(1)}|Z^{(1)},Y^{(2)})\big|\Big]\label{eq: tmp1}
\end{align}
On the other hand, for some values $z\in \reals^{d+1},y\in \reals $ consider the following measurable set:
\[
A_{z,y}=\{x\in \reals^d: T(x,y)\leq T(z) \}\,.
\]
By using this definition of set $A_{z,y}$ in \eqref{eq: tmp1} and shorthands $W=(Z^{(1)},Y^{(2)})$ we arrive at 
\begin{align}
\Big|\pi-\frac{1}{2}\Big|&\leq \E\Big[\big|\prob(X^{2}_{\swap(i,j)}\in {A}_W |W)-\prob(X^{(2)} \in A_W|W)\big|\Big]\nonumber \\
&\leq \E\Big[d_\tv\big( p_{X^{(2)}_{\swap(i,j)}|W}, p_{X^{(2)}|W} \big) \Big]\,,\label{eq: tmp2}
\end{align}
where the last inequality follows the definition of the total variation distance. Since $Z^{(1)}$ and $Z^{(2)}$ are independent random variables, we get that 
\begin{align*}
d_\tv\Big( p_{X^{(2)}_{\swap(i,j)}|W}, p_{X^{(2)}|W} \Big) &=d_\tv\Big( p_{X^{(2)}_{\swap(i,j)}|Y^{(2)}}, p_{X^{(2)}|Y^{(2)}} \Big)\\
&=d_\tv\big( p_{X_{\swap(i,j)}|Y}, p_{X|Y} \big)\,,
\end{align*}
where the last relation comes from the fact that random variable $(x,y)\sim p_{x,y}$ and $(x^{(2)},y^{(2)})$ has a similar density function.  Using the above relation in \eqref{eq: tmp2} yields
\begin{align*}
\left |\pi-\frac{1}{2}\right| &\leq \E\left[d_\tv\Big( p_{X^{}_{\swap(i,j)}|Y^{}}, p_{X^{}|Y^{}} \Big) \right]\\
&=d_\tv\left( p_{X_{\swap(i,j)},Y}, p_{X,Y} \right)\,.
\end{align*}
In the next step, for $x\in \reals^d$ and $y\in \reals$ let $p(x,y)$ and $q(x,y)$ respectively denote the density functions of $(X_{\swap(i,j)},Y)$ and $(X,Y)$. From the above relation we get
\begin{align*}
\left |\pi-\frac{1}{2}\right| & \leq \frac{1}{2}\int \left|p(x,y)-q(x,y) \right| \de x \de y\,.
\end{align*}
On the other hand, by rewriting the total variation distance of the joint random variables get
\begin{align*}
|p(x,y)-q(x,y)|&=|p(x)p(y|x)-q(x)q(y|x)|\nonumber\\
&=|p(x)p(y|x)-p(x)q(y|x)\\
&\, +p(x)q(y|x)-q(x)q(y|x)|\nonumber\\
&\leq p(x)|p(y|x)-q(y|x)|\\
&\,+ |p(x)-q(x)|q(y|x)\,.
\end{align*}
Plugging this into the above relation yields
\begin{align*}
\left |\pi-\frac{1}{2}\right| &\leq \frac{1}{2}\int p(x)|p(y|x)-q(y|x)|\de x \de y\\
&\,+\frac{1}{2}\int |p(x)-q(x)|q(y|x)\de x\de y\,.
\end{align*}
In the next step, by integration with respect to $y$ we get
\begin{align*}
\left |\pi-\frac{1}{2}\right|&\leq \frac{1}{2}\int p(x)|p(y|x)-q(y|x)|\de x \de y\\
&\,+\frac{1}{2}\int |p(x)-q(x)|\de x\,.
\end{align*}
This implies that
\begin{align*}
\left |\pi-\frac{1}{2}\right|&\leq \E_X[d_{\tv}(p_{Y|X_{\swap(i,j)}},p_{Y|X})]+d_\tv(p_X,p_{X_{\swap(i,j)}})\,.
\end{align*}
Finally, under the null hypothesis \ref{eq: null} and the fact that 
$\tau_X\geq d_\tv( p_X, p_{X_{\swap(i,j)}})$ we get
\begin{equation}\label{eq: deviation-bound}
\left|\pi-\frac{1}{2}\right| \leq \tau_{X}+\tau\,.
\end{equation}
 Any deviation from this range is accounted as evidence against the null hypothesis \ref{eq: null}. In Algorithm \ref{algorithm: importance}, for each $1\leq m\leq n/2$, it is easy to observe that each random variable $\ind\left(T(X^{(m)},Y^{(m)})\geq T(\tX^{(m)},\widetilde{Y}^{(m)})  \right)$ is a Bernoulli with success probability $\pi$.    
In the next step, by an application of Hoeffding's inequality for every $t\geq 0$ and sum of $n/2$ independent Bernoulli random variables we get
\[
\prob\left(\Big|\frac{2}{n}\sum\limits_{i=1}^{n/2}\ind\big(T(x_i,y_i)\leq T(\tx_i,\ty_i)\big) -\pi \Big|\geq t \right) \leq 2\exp(-nt^2)\,.
\]
Therefore, for statistics $U_n$ as per Algorithm \ref{algorithm: importance} we get  

\begin{equation}\label{eq: hoefding}
\prob(|U_n -\pi |\geq t) \leq 2\exp(-nt^2)\,,\quad \forall t\geq 0
\end{equation}
We next consider $\delta\geq \tau+\tau_X$ and use triangle inequality to obtain
\begin{align*}
    \prob\left(\Big|U_n-\frac{1}{2}\Big|\geq \delta\right)&\leq   \prob\left(\Big|U_n-\pi\Big|+\Big|\pi-\frac{1}{2}\Big|\geq \delta\right)\\
    &\leq \prob(|U_n-\pi|\geq \delta-\tau-\tau_X)\\
    &\leq 2\exp(-n(\delta-\tau-\tau_X)^2)\,.
\end{align*}
Where in the penultimate relation we used \eqref{eq: deviation-bound}, and the last relation follows \eqref{eq: hoefding}. By letting $\alpha=\delta-\tau-\tau_X$, we get 
\[
\prob\left(\Big|U_n-\frac{1}{2}\Big|\geq \tau+\tau_X+\sqrt{\frac{\log \frac{2}{\alpha}}{n}}\right)\leq \alpha\,.
\]
This completes the proof.


\subsection{Proof of Proposition \ref{propo: glm}}\label{sec: propo: glm}
We start with $\th_i=\th_j$, and we want to show that the symmetric influence property holds.  We have 
\begin{align*}
    p_{Y|X_{\swap(i,j)}}(y|x)&=p_{Y|X}(y|x_{\swap(i,j)})\\
    &=\prob(Y=1|X=x_{\swap(i,j)})\\
    &=\left(1+\exp(-x_{\swap(i,j)}^\sT\beta)\right)^{-1}\\
&=\Big(1+\exp\big(-\beta_ix_j-\beta_jx_i-\sum_{\ell \neq i,j}x_\ell\beta_\ell\big)\Big)^{-1}\,.
\end{align*}
Using $\beta_i=\beta_j$ yields
\begin{align*}
p_{Y|X_{\swap(i,j}}(y|x)&=\Big(1+\exp\big(-\beta_jx_j-\beta_ix_i-\sum_{\ell \neq i,j}x_\ell\beta_\ell\big)\Big)^{-1}\\
&=\Big(1+\exp\big(-\sum_{\ell}x_\ell\beta_\ell\big)\Big)^{-1}\\
&=p_{Y|X}(y|x)\,.
\end{align*}
This completes the proof for the first part.  For the other direction, suppose that the symmetric influence for $i, j$ holds, thereby for every $x\in \reals^d$ we have
\[
 \prob(Y=+1|X_{\swap(i,j)}=x)= \prob(Y=+1|X=x)\,.
\]
By using $p_{Y|X_{\swap(i,j)}}(y|x)=p_{Y|X}(y|x_{\swap(i,j)})$ along with the logistic regression relation, we get
\begin{align*}
&\Big(1+\exp\big(-\beta_ix_j-\beta_jx_i-\sum_{\ell \neq i,j}x_\ell\beta_\ell\big)\Big)^{-1}\\
&=\Big(1+\exp\big(-\sum_{\ell}x_\ell\beta_\ell\big)\Big)^{-1}\,.
\end{align*}
In the next step, using the function $\log(\frac{u}{1-u})$ on the both sides, we get
\[
\beta_i x_i +\beta_j x_j=\beta_i x_j + \beta_j x_i \,.
\]
Since this must hold for all $x_i,x_j$ values, we must have $\beta_i=\beta_j$.
The proof for the linear regression setting follows the exact similar argument.

\subsection{Proof of Proposition \ref{propo: gaussian-swap}}
Since $x$ is a multivariate Gaussian, it means that its coordinates are jointly Gaussian random variables, therefore swapping the location of two coordinates $i$ and $j$ does not change the joint Gaussian property. On the other hand, from the linear transform $x_{\swap(i,j)}=P_{ij}x$ it is easy to arrive at $x_{\swap(i,j)}\sim \normal(P_{ij}\mu,P_{ij}\Sigma P_{ij})$. We are only left with upper bounding the KL divergence of density functions $\normal(\mu,\Sigma)$ and $\normal(P_{ij}\mu,P_{ij}\Sigma P_{ij})$. For this end, we borrow a result from \cite{duchi2007derivations} for kl-divergence of multivariate Gaussian distributions. Formally we have,
\begin{align*}
&d_{\mathsf{kl}}\left(\mathcal{N}(\mu_1,\Sigma_1)\|\mathcal{N}(\mu_2,\Sigma_2)\right)\\
&=\frac{1}{2}\left(\log \frac{\det{\Sigma_2}}{\det{\Sigma_1}}-d+\trace(\Sigma_2^{-1}\Sigma_1)+(\mu_2-\mu_1)^\sT\Sigma_2^{-1}(\mu_2-\mu_1) \right)\,.
\end{align*}
By replacing $\Sigma_2=\Sigma$ and $\Sigma_1= P_{ij}\Sigma P_{ij}$ along with the fact that $\det P_{ij}=-1$, we arrive at
\begin{align*}
&d_{\kl}\left(\cL(X_{\swap(i,j)})\|\cL(X)\right)=\frac{1}{2}\left(-d+\trace(\Sigma^{-1}P_{ij}\Sigma P_{ij})+(\mu-P_{ij}\mu)^\sT\Sigma^{-1}(\mu-P_{ij}\mu) \right)\,.
\end{align*}
Finally using Pinsker's inequality\footnote{ For two denisty functions $p,q$ this holds $d_{\tv}(p,q)\leq \sqrt{\frac{d_{\mathsf{kl}}(p\|q)}{2}}$} completes the proof. 

\subsection{Proof of Proposition \ref{propo: unif-binary}}
In this setup, form the construction of the feature vector $x\in \{0,1\}^d$ it is easy to get that for every $\alpha\in \{0,1\}^d$ we have
\[
\prob(x=\alpha)=\frac{\ind(|\alpha|=m)}{\binom{d}{m}}\,.
\]
From this structure, since swapping the coordinates does not change the number of non-zero entries of the binary feature vector, we get $|\alpha|=|\alpha_{\swap(i,j)}|$. Thereby, we get
\[
\prob(x=\alpha)=\prob(x_{\swap(i,j)}=\alpha)\,, \quad \forall \alpha\in \{0,1\}^d. 
\]

Therefore $d_{\tv}\left(\cL(X),\cL(X_{\swap(i,j)})\right)=0$.

\subsection{Proof of Theorem \ref{thm: power}}
Let
\[
\pi= \prob\left(T(X^{(1)},Y^{(1)})\geq T(X^{(2)}_{\swap(i,j)},Y^{(2)} \right)\,.
\]
For the sake of simplicity, we adopt the following shorthands: $T_1=T(X^{(1)},Y^{(1)})$ and $T_2=T(X^{(2)}_{\swap(i,j)},Y^{(2)})$. This gives us
\begin{align*}
\pi&=\prob(T_1\geq T_2)\\
&=\E_{T_1}\left[\prob(T_1\geq T_2|T_1) \right]\\
&=\E_{T_1}\left[G_{T}(T_1)\right]\\
&=\int G_T(t)\de F_T(t)\\
&=\int\limits_{0}^{1} G_T(F_T^{-1}(u))\de u\,.
\end{align*}
In  the next step, we let $\delta=2\exp(-n\beta^2)$, then by plugging this relation in the given condition in Theorem \ref{thm: power} we arrive at
\begin{equation}\label{eq: lower-pi}
|\pi-1/2|\geq \delta + \tau+\tau_X+\sqrt{\frac{\log(2/\alpha)}{n}}\,.
\end{equation}
We now focus on the decision rule \eqref{eq: decision}. Let $\tau'=\tau+\tau_X$, then we get
\begin{align}\label{eq: decision-3}
\prob(\Psi(\bX,\bY)=1)&=\prob\left(|U_n-1/2|\geq \tau'+\sqrt{\frac{\log(2/\alpha)}{n}}\right)\,.
\end{align}
On the other hand, from triangle inequality we have $|U_n-1/2|\geq |\pi-1/2|-|U_n-\pi|$. Plugging this into \eqref{eq: lower-pi} yields
\[
|U_n-1/2| \geq \delta + \tau'+\sqrt{\frac{\log(2/\alpha)}{n}}-|U_n-\pi|
\]
Combining this with \eqref{eq: decision-3} gives us
\begin{align}\label{eq: decision-2}
\prob(\Psi(\bX,\bY)=1)&\geq\prob\left(\delta\geq |U_n-\pi| \right)\nonumber\\
&=1-\prob\left(\delta\leq |U_n-\pi| \right)\,.
\end{align}

In the next step, we return to the given relation for $U_n$ in Algorithm \ref{algorithm: importance}. From the definition of $\pi$, for each $m$ we have
$$\prob\left(T\big(X^{(m)},Y^{(m)}\big)\leq T\big(\tX^{(m)},\widetilde{Y}^{(m)}\big)\right)=\pi\,.$$Therefore by an application of the Hoeffding's inequality we get
\begin{align*}
\prob\left( |U_n-\pi| \geq \delta \right)\leq \sqrt{\frac{\log(2/\delta)}{n}}\,.
\end{align*}
Finally, recalling $\delta=2\exp(-n\beta^2)$ yields
\begin{align*}
\prob\left( |U_n-\pi| \geq \delta \right)\leq \beta \,.
\end{align*}
Using this in \eqref{eq: decision-2} completes the proof. In this case, statistical power not smaller than $1-\beta$ can be achieved.

\subsection{Proof of Theorem \ref{thm: power-linear}}
From the isotropic Gaussian distribution, we have $\tau_X=0$. We next start by the ODC function $G_T\circ F_T^{-1}$. For this end, we start by the definition of $F_T$ where for some non-negative $t$ we have:
\begin{align*}
F_T(t)&=\prob(|Y^{(1)}-\hth^\sT X^{(1)}|\leq t)\\
&=\prob(|(\th^*-\hth)^\sT X^{(1)}+\eps_1|\leq t)\\
&=\prob(-t\leq (\th^*-\hth)^\sT X^{(1)}+\eps_1\leq t)\,.
\end{align*}
On the other hand, we know that $x^\sT (\hth-\th^*)+\eps$ has a Gaussian  distribution $\normal(0,\|\th^*-\hth\|_2^2+\sigma^2)$. This brings us
\begin{align}
F_T(t)&=\prob(-t\leq (\th^*-\hth)^\sT X^{(1)}+\eps_1\leq t)\nonumber\\
&=\prob\left(\frac{-t}{\sqrt{\|\th^*-\hth\|_2^2+\sigma^2}}\leq \frac{(\th^*-\hth)^\sT X^{(1)}+\eps_1}{\sqrt{\|\hth-\th^*\|_2^2+\sigma^2}}\leq \frac{t}{\sqrt{\|\hth-\th^*\|_2^2+\sigma^2}}\right)\nonumber\\
&=\Phi\left(\frac{t}{\sqrt{\sigma^2+\|\th^*-\hth\|_2^2}} \right)-\Phi\left(-\frac{t}{\sqrt{\sigma^2+\|\th^*-\hth\|_2^2}} \right)\nonumber\\
&=2\Phi\left(\frac{t}{\sqrt{\sigma^2+\|\th^*-\hth\|_2^2}} \right)-1\label{eq: linear-F}\,,
\end{align}
where the last line comes from the fact that $\Phi(t)+\Phi(-t)=1$ for every real value $t$. We introduce the shorthand $\hth_\swap=\hth_{\swap(i,j)}$, then by a similar argument we get
\begin{align}
G_T(t)&=\prob(|Y^{(2)}-\hth^\sT X^{(2)}_{\swap(i,j)}|\leq t)\nonumber\\
&=\prob(|(\th^*-\hth_{\swap})^\sT X^{(2)}+\eps_2|\leq t)\nonumber\\
&=2\Phi\left(\frac{t}{\sqrt{\sigma^2+\|\th^*-\hth_\swap\|_2^2}} \right)-1\label{eq: linear-G}\,,
\end{align}
By combining \eqref{eq: linear-F} and \eqref{eq: linear-G}  we get 
\[
F_T\circ G_T^{-1}(u)=2\Phi\left( \frac{\sigma_2}{\sigma_1}\Phi^{-1}\Big(\frac{u+1}{2}\Big) \right)-1\,,
\]
for $\sigma_2$ and $\sigma_1$ given by
\[
\sigma_1^2=\sigma^2+\|\th^*-\hth\|_2^2\,,\quad \sigma_2^2=\sigma^2+\|\th^*-\hth_\swap\|_2^2\,.
\]
We consider $\gamma=\frac{\sigma_2}{\sigma_1}$. Plugging this into the power expression in Theorem \ref{thm: power} we arrive at
\[
F_T(G_T^{-1}(u))-u =2\left[ \Phi \bigg(\gamma \Phi^{-1}\Big(\frac{u+1}{2}\Big)\bigg)-\frac{u+1}{2} \right]\,.
\]
In the next step, by using the change of variable $v=\frac{u+1}{2}$ we get
\[
\int_0^1[F_T(G_T^{-1}(u))-u]\de u = 4\int_{\frac{1}{2}}^1\left[ \Phi \big(\gamma \Phi^{-1}(v)\big)-v \right]\de v\,.
\]
We then introduce function $\psi:[0,+\infty]\rightarrow \reals$ as following
\[
\psi(\gamma)=4\int_{\frac{1}{2}}^1 \Phi \Big(\gamma \Phi^{-1}(v)\Big)\de v\,.
\]
This implies that
\begin{equation}\label{eq: psi-gamma}
\psi(\gamma)-\psi(1)=\int_0^1[F_T(G_T^{-1}(u))-u]\de u\,.
\end{equation}
By differentiating $\psi(.)$ with respect to $\gamma$ in its original definition we obtain
\begin{align*}
    \frac{\de\psi}{\de \gamma}&=4\frac{\partial}{\partial \gamma} \int_{\frac{1}{2}}^1 \Phi \Big(\gamma \Phi^{-1}(v)\Big)\de v\\
    &=4\int_{\frac{1}{2}}^1  \Phi^{-1}(v) \varphi \Big(\gamma \Phi^{-1}(v)\Big)\de v\,.
    \end{align*}
We next use $s=\Phi^{-1}(v)$ to arrive at the following   
    \begin{align*}
   \frac{\de\psi}{\de \gamma} &=4\int_{0}^{+\infty}  s \varphi \Big(\gamma s\Big) \varphi(s)\de s\\
    &=\frac{4}{2\pi}\int_{0}^{+\infty}  s \exp\Big(-\frac{s^2}{2}(1+\gamma^2)\Big)\de s\\
    &=\frac{2}{\pi(\gamma^2+1)}\int_0^{+\infty} s\exp(-s^2/2)\de s\\
    &=\frac{2}{\pi(\gamma^2+1)}\,.
\end{align*}
Since the differentiation of $\psi$ with respect to $\gamma$ is provided above, we then can use this and obtain the closed form equation for $\psi(u)$. This indeed is given by
\[
\psi(\gamma)=C+\frac{2}{\pi}\arctan(\gamma)\,,
\]
For some constant value $C$. In order to find $C$, note that $\psi(1)=4\int\limits_{\frac{1}{2}}^{1}v\de v=\frac{3}{2}$. This brings us $\psi(\gamma)=1+\frac{2}{\pi}\arctan(\gamma)$. Using this in \eqref{eq: psi-gamma} yields
\begin{align*}
\left|\int_0^1[F_T(G_T^{-1}(u))-u]\de u\right|&=|\psi(\gamma)-\psi(1)|\\
&=\frac{2}{\pi}\left|\arctan(\gamma)-\arctan(1)\right|\,.
\end{align*}
On the other hand, from the identity $\arctan(x)-\arctan(y)=\arctan\frac{x-y}{1+xy}$ we arrive at:
\begin{align*}
\left|\int_0^1[F_T(G_T^{-1}(u))-u]\de u\right|&=\frac{2}{\pi}\left|\arctan\Big(\frac{\gamma-1}{1+\gamma}\Big)\right|\\
&=\frac{2}{\pi}\arctan\Big(\frac{|\gamma-1|}{1+\gamma}\Big)\,,
\end{align*}
where in the last relation we used $\arctan(|.|)=|\arctan(.)|$ (note that $\gamma\geq 0$).
We next use $\gamma=\sigma_2/\sigma_1$ to get
\begin{equation}\label{eq: linear-atan}
\left|\int_0^1[F_T(G_T^{-1}(u))-u]\de u\right|=\frac{2}{\pi}\arctan\left(\frac{|\sigma_1-\sigma_2|}{\sigma_1+\sigma_2}\right)\,.
\end{equation}
On the other hand, from $\sigma_1^2+\sigma_2^2\geq 2\sigma_1\sigma_2$ we get
\[
\Delta_T=\frac{|\sigma_1-\sigma_2|}{|\sigma_1+\sigma_2|}\geq \frac{|\sigma_1^2-\sigma_2^2|}{2(\sigma_1^2+\sigma_2^2)}
\] 
We then use this with the definition of $\sigma_1,\sigma_2$ to get
\begin{align*}
\Delta_T&\geq \frac{1}{2}\frac{\left|\|\th^*-\hth\|_2^2-\|\th^*-\hth_\swap\|_2^2\right|}{2\sigma^2+\|\th^*-\hth\|_2^2+\|\th^*-\hth_\swap\|_2^2}\\
&=\frac{1}{2}\frac{\left|-2\hth^\sT\th^*+2\hth^\sT_\swap \th^*\right|}{2\sigma^2+2\|\th^*\|_2^2+2\|\hth\|_2^2-2\hth^\sT\th^*-2\hth_\swap^\sT\th^*}\,,
\end{align*}
where we used $\|\th\|_2=\|\th_{\swap}\|_2$. In the next step, since $\hth_{\swap,\ell}=\hth_{\ell}$ for all $\ell\neq i, j$ we get
\begin{align*}
\Delta_T&\geq\frac{1}{2}\frac{\left|-\hth_i\th^*_i-\hth_j\th^*_j+\hth_i\th^*_j+\hth_j\th^*_i \right|  }{\sigma^2+\|\th^*\|_2^2+\|\hth\|_2^2-\hth^\sT\th^*-\hth_\swap^\sT\th^*}\\
&=\frac{1}{2}\frac{\left|-\hth_i\th^*_i-\hth_j\th^*_j+\hth_i\th^*_j+\hth_j\th^*_i \right|  }{\sigma^2+\|\th^*-\hth\|_2^2+\hth^\sT\th^*-\hth_\swap^\sT\th^*}
\end{align*}
In the next step, by using the observation that $\hth_{\swap,\ell}=\hth_{\ell}$ for all $\ell\neq i, j$ another time we get

\begin{align*}
\Delta_T&\geq \frac{1}{2}\frac{|\hth_i-\hth_i||\th^*_i-\th^*_j| }{\sigma^2+\|\th^*-\hth\|^2+(\th^*_i-\th^*_j)(\hth_i-\hth_j)}
\end{align*}
Thereby we get
\begin{align*}
\Delta_T &\geq \frac{1}{2}\frac{|\hth_i-\hth_i||\th^*_i-\th^*_j| }{\sigma^2+\|\th^*-\hth\|^2+|\th^*_i-\th^*_j||\hth_i-\hth_j|}
\end{align*}
Using the above relation in \eqref{eq: linear-atan} we get
\begin{equation}\label{eq: linear-atan-2}
\left|\int_0^1[F_T(G_T^{-1}(u))-u]\de u\right|\geq \frac{2}{\pi}\arctan\left( \frac{1}{2}\frac{|\hth_i-\hth_i||\th^*_i-\th^*_j| }{\sigma^2+\|\th^*-\hth\|^2+|\th^*_i-\th^*_j||\hth_i-\hth_j|} \right)
\end{equation}

By recalling the given condition in Theorem \ref{thm: power-linear} we have
\[
|\th^*_i-\th^*_j|\geq \frac{2\tan(\frac{\pi}{2}(\rho_n(\alpha,\beta,\tau)))}{1-2\tan(\frac{\pi}{2}(\rho_n(\alpha,\beta,\tau))) }\frac{\Big(\sigma^2+\|\hth-\th^*\|_2^2 \Big)}{|\hth_i-\hth_j|}\,,
\]
By using $\tan(\frac{\pi}{2}(\rho_n(\alpha,\beta,\tau)))\leq \frac{1}{2}$ in the above relation we get
\begin{equation}\label{eq: linear-atan-3}
\frac{2}{\pi}\arctan\left( \frac{1}{2}\frac{|\hth_i-\hth_i||\th^*_i-\th^*_j| }{\sigma^2+\|\th^*-\hth\|^2+|\th^*_i-\th^*_j||\hth_i-\hth_j|} \right)\geq \rho_n(\alpha,\beta,\tau).
\end{equation}
By combining \eqref{eq: linear-atan-2} and \eqref{eq: linear-atan-3} we get
\[
\left|\int_0^1[F_T(G_T^{-1}(u))-u]\de u\right|\geq \rho_n(\alpha,\beta,\tau)\,.
\]
Finally using Theorem \ref{thm: power-linear} completes the proof, 

\subsection{Proof of Theorem \ref{thm: power-binary}}
We first show that in this case, ($\tau=0$) for mixture of Gaussians, under the null hypothesis, we have $\tau_X=0$. For this end, from the Bayes' formula it is easy to get $\cL(Y|X)=\mathsf{Bern}(g(x,\mu))$ with
\[
g(x,\mu)=\frac{1}{1+\frac{1-q}{q}e^{-x^\sT \mu}}\,.
\]
With a similar argument, it can be observed that
\[
\cL(Y|X)=\mathsf{Bern}(g(x,\mu_{\swap(i,j)}))\,.
\]

Given that $d_{\tv}(\mathsf{Bern}(a),\mathsf{Bern}(b))=|a-b|$, under the null hypothesis (with $\tau=0$) we must have $g(x,\mu)=g(x,\mu_{\swap(i,j)})$ almost surely for all $x$ values. This implies that $x^\sT \mu=x^\sT \mu_{\swap(i,j)}$ almost surely, thereby we have $\mu_i=\mu_j$. In the next step, we show that if $\mu_i=\mu_j$ then $\tau_X=0$. We then note that

\begin{align*}
\cL(X)&=q \normal(+\mu,I_d)+ (1-q) \normal (-\mu, I_d)\,, \\
\cL(X_{\swap(i,j)})&=q \normal(+\mu_{\swap(i,j)},I_d)+ (1-q) \normal (-\mu_{\swap(i,j)}, I_d)\,. 
\end{align*}

In the next step, using $\mu_i=\mu_j$ we realize that $\mu_{\swap(i,j)}=\mu$, therefore $\cL(X)=\cL(X_{\swap(i,j)})$. This implies that $\tau_X=0$.

For the rest of the proof, we follow a similar argument as per proof of Theorem \ref{thm: power-linear} and we first characterize cdf functions $F_T$ and $G_T$. In this case we have
\begin{align*}
F_T(t)&=\prob(Y^{(1)}\hth^\sT X^{(1)}\leq t)\\
&=q\prob(\hth^\sT X^{(1)}\leq t|Y^{(1)}=+1)+\\
&\,+(1-q)\prob(-\hth^\sT X^{(1)}\leq t|Y^{(1)}=-1)\\
&=q\prob(Z^{+}\leq t)+(1-q)\prob(Z^{-}\leq t)\,,
\end{align*}
where $Z_+\sim \normal(\mu^\sT\hth,\|\hth\|_2^2)$ and $Z_-\sim \normal(-\mu^\sT\hth,\|\hth\|_2^2)$. This yields
\begin{align*}
  F_T(t)&=q\Phi\Big(\frac{t-\hth^\sT\mu}{\|\hth\|_2}\Big) +(1-q)\left(1-\Phi \Big(\frac{-t+\hth^\sT\mu}{\|\hth\|_2}\Big) \right)\\
  &=\Phi\left(\frac{t-\hth^\sT\mu}{\|\hth\|_2}\right)\,,
\end{align*}
where in the last line we used $\Phi(t)+\Phi(-t)=1$. 
We next introduce the shorthands $\hth_{\swap}=\hth_{\swap(i,j)}$ and $\mu_\swap=\mu_{\swap(i,j)}$, then by a similar argument we arrive at
\[
G_T(t)=\Phi\left(\frac{t-\hth_{\swap}^\sT\mu}{\|\hth_\swap\|_2}\right)
\]
Since $\hth_{\swap}^\sT\mu=\mu_{\swap}^\sT \hth$ and $\|\hth_\swap\|=\|\hth\|$ the expression for $G_T(t)$ can be written as the following:
\[
G_T(t)=\Phi\left(\frac{t-\hth^\sT\mu_{\swap}}{\|\hth\|_2}\right)
\]
In the next step, it is easy to compute the quantile function  $G_T^{-1}(u)=\|\hth\|_2 \Phi^{-1}(u)+\hth^\sT \mu_{\swap}$. This brings us
\[
F_T(G_T^{-1}(u))=\Phi\left(\Phi^{-1}(u)+\frac{\hth^\sT(\mu_{\swap}-\mu)}{\|\hth\|_2}\right)\,.
\]
By introducing $\lambda=\frac{\hth^\sT(\mu_{\swap}-\mu)}{\|\hth\|_2}$ and the function  $\rho(\lambda)=\int_0^1 \Phi(\Phi^{-1}(u)+\lambda)\de u$ we obtain
\[
\int_0^1 F_T(G_T^{-1}(u))\de u =\rho(\lambda)\,.
\]
On the other hand, by differentiating $\rho(\lambda)$ with respect to $\lambda$  we get
\begin{align*}
\frac{\partial \rho}{\partial \lambda}&=\frac{\partial}{\partial \lambda} \int_0^1 \Phi(\Phi^{-1}(u)+\lambda)\de u\\
&=\int_0^1 \varphi(\Phi^{-1}(u)+\lambda)\de u\,.
\end{align*}
In the next step,  by using the change of variable $s=\Phi^{-1}(u)$ we get that
\begin{align*}
\frac{\partial \rho}{\partial \lambda}&=\int_{-\infty}^{\infty} \varphi(s+\lambda)\varphi(s)\de s\\
&=\frac{1}{2\pi} \int_{-\infty}^{\infty} \exp\left(-\frac{(s+\lambda)^2}{2}-\frac{s^2}{2} \right)\de s\\
&=\frac{\exp(-\lambda^2/4)}{2\pi}\int_{-\infty}^{+\infty} \exp\left(-\frac{(\sqrt{2}s+\lambda/\sqrt{2})^2}{2}\right)\de s\\
&=\frac{\exp(-\lambda^2/4)}{2\sqrt{2}\pi}\int_{-\infty}^{+\infty} \exp\left(-\frac{(t+\lambda/\sqrt{2})^2}{2}\right)\de t=\frac{\exp(-\lambda^2/4)}{2\sqrt{\pi}}\,.
\end{align*}

Therefore we get $\rho(\lambda)=\rho(0)+\int_0^{\lambda}\frac{\exp(-s^2/4)}{2\sqrt{\pi}}=\rho(0)+\Phi(\frac{\lambda}{\sqrt{2}})-\frac{1}{2}$, Since $\rho(0)=1/2$, we arrive at $\rho(\lambda)=\Phi\left(\frac{\lambda}{\sqrt{2}}\right)$. Next from the definition of $\rho(\lambda)$ we have
\[
\int_0^1\Big[F_T(G_T^{-1}(u))-u\Big]\de u=\rho(\lambda)-\rho(0)\,.
\]
In the next step, we use the equivalent value of $\lambda$ in the function $\rho(\lambda)$ to get

\[
\int_0^1\Big[F_T(G_T^{-1}(u))-u\Big]\de u=\Phi\left(\frac{\hth^\sT(\mu_{\swap}-\mu)}{\sqrt{2}\|\hth\|_2} \right)-\Phi(0)\,.
\]
Therefore we get
\[
\left|\int_0^1\Big[F_T(G_T^{-1}(u))-u\Big]\de u \right|=\left|\Phi\left(\frac{\hth^\sT(\mu_{\swap}-\mu)}{\sqrt{2}\|\hth\|_2} \right)-\Phi(0)\right|\,.
\]
On the other hand, the normal cdf satisfies the following property
\[
\left|\Phi(t)-\frac{1}{2}\right| = \Phi(|t|)-\frac{1}{2}\,, \forall t\in\reals 
\]
By using this we get
\begin{equation}\label{eq: binary-Phi}
\left|\int_0^1\Big[F_T(G_T^{-1}(u))-u\Big]\de u \right|= \Phi\left(\bigg|\frac{\hth^\sT(\mu_{\swap}-\mu)}{\sqrt{2}\|\hth\|_2}\bigg| \right)-\frac{1}{2}\,.
\end{equation}
In the next step, by using the fact that $\mu_{\swap,\ell}=\mu_\ell$ for $\ell\neq i,j$ we get that
\begin{align*}
\hth^\sT(\mu_{\swap}-\mu)&=\hth_i(\mu_{\swap,i}-\mu_i)+\hth_j(\mu_{\swap,j}-\mu_j)\\
&=\hth_i(\mu_{j}-\mu_i)+\hth_j(\mu_{i}-\mu_j)\\
&=-(\hth_i-\hth_j)(\mu_i-\mu_j)\,.
\end{align*}
Using this in \eqref{eq: binary-Phi} yields
\begin{equation}\label{eq: binary-Phi2}
\left|\int_0^1\Big[F_T(G_T^{-1}(u))-u\Big]\de u \right|= \Phi\left(\frac{|(\hth_i-\hth_j)(\mu_i-\mu_j)|}{\sqrt{2}\|\hth\|_2} \right)-\frac{1}{2}
\end{equation}

On the other hand, by recalling the condition on $|\mu_i-\mu_j|$ from Theorem \ref{thm: power-binary} we have
\begin{equation}\label{eq: binary-Phi3}
|\mu_i-\mu_j|\geq \Phi^{-1}\left(\rho_n(\alpha,\beta,0)+2\Phi\left(\frac{|\mu_i-\mu_j|}{\sqrt{2}} \right)-\frac{1}{2} \right)\frac{\sqrt{2}\|\hth\|_2}{|\hth_i-\hth_j|}
\end{equation}
Combining \eqref{eq: binary-Phi2} and \eqref{eq: binary-Phi3} yields
\[
\left|\int_0^1\Big[F_T(G_T^{-1}(u))-u\Big]\de u \right| \geq \rho_n(\alpha,\beta,0)\,.
\]

Finally, using Theorem \ref{thm: power} completes the proof. 

\section{Additional Numerical Experiments}
\subsection{Size of the test (full experiments)} 
We refer to Figure \ref{fig: isotropic-appx} for experiment on the size of the test. 

{
\begin{figure*}[h!]
\centering
\subfigure[$\alpha=0.1$]{ 
\includegraphics[width=0.3\linewidth]{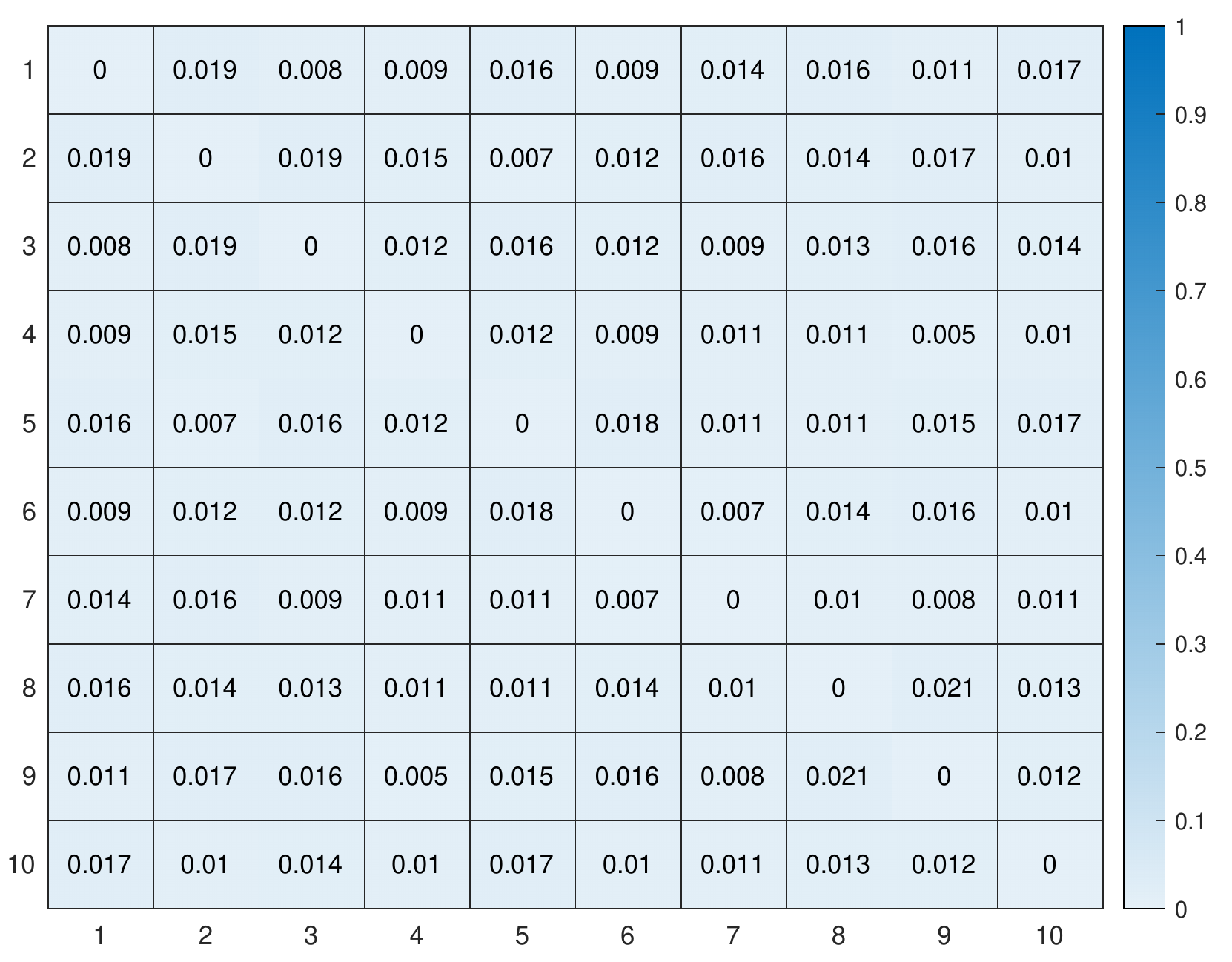}
}
\hfill
\subfigure[$\alpha=0.15$]{
\includegraphics[width=0.3\linewidth]{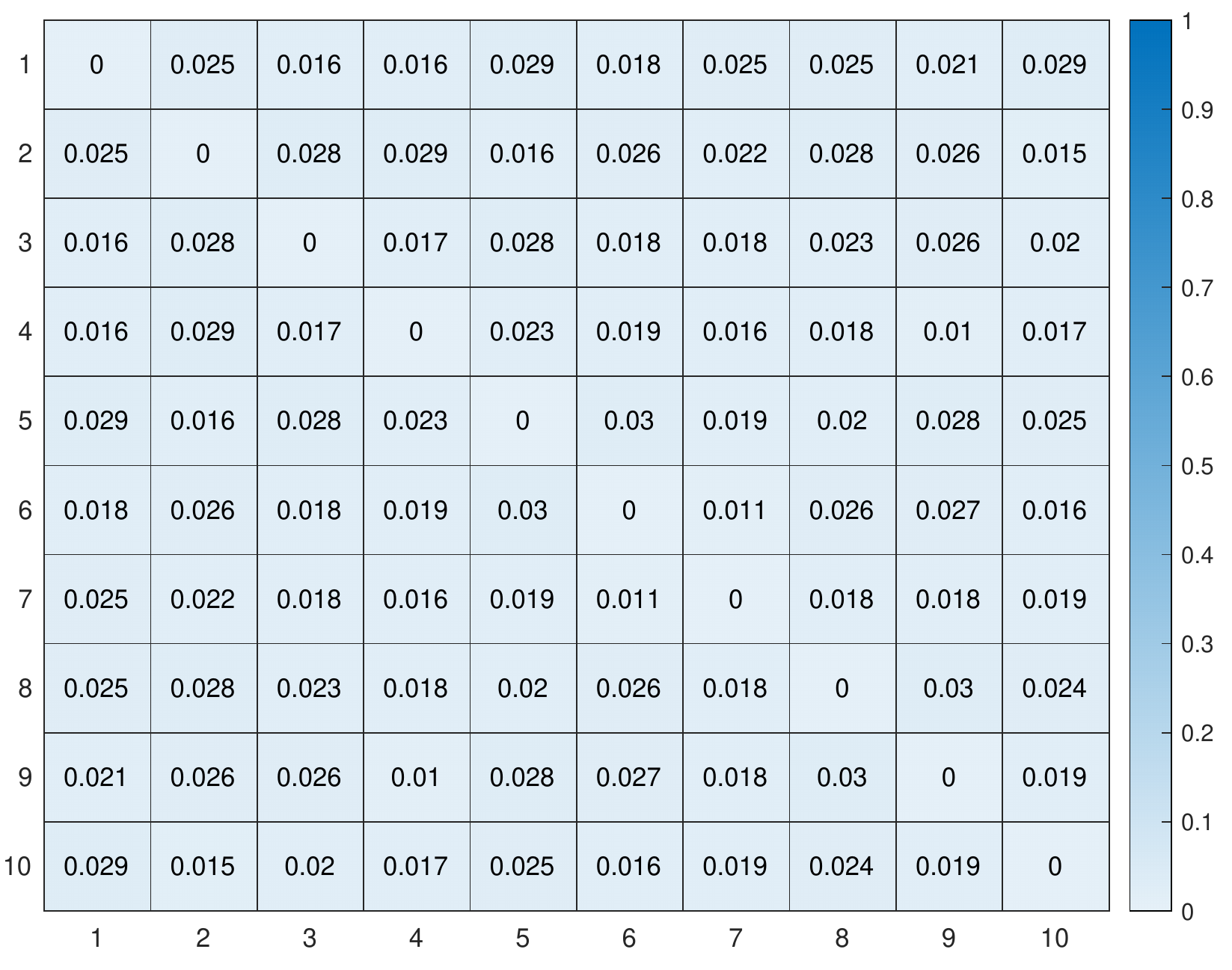}
}
\hfill
\subfigure[$\alpha=0.2$]{
\includegraphics[width=0.3\linewidth]{isotropic_size_20.pdf}
}
\hspace{-2mm}

\vspace{-2mm}
\caption{Average rejection rate of the null hypothesis \eqref{eq: null} for $\tau=0$ and features coming from an isotropic Gaussian distribution $x\sim \normal(0,I_{10})$. In this experiment, we consider $y|x \sim \normal(x^\sT S x,1)$ for a positive definite matrix $S_{i,j}=1+\ind(i=j)$ ($2$ on diagonal and $1$ on off-diagonal entries). The structure of $S$ implies that the symmetric influence holds for every pair of features. We consider three significance levels $\alpha=0.1,0.15,0.2$ (from left to right). The small cell $(i,j)$ in each plot, represents rejection rates for testing symmetric influence for features $i$ and $j$. In this experiment, the number of data points is $1000$ and the method is run with the score function $T(x,y)=|y-x^\sT \hth|$ for $\hth\sim\normal(0,I_{10})$. The reported numbers are averaged over $1000$ experiments. It can be seen that the size of the test is controlled at the pre-determined significance levels.  }\label{fig: isotropic-appx}

\end{figure*} 
}

\subsection{Power of the test (full experiments)}
\begin{figure*}[t]
\centering
\subfigure[$\sigma=1$]{ 
\includegraphics[width=0.3\linewidth]{isotropic_rho_1.pdf}
}
\hfill
\subfigure[$\sigma=2$]{
\includegraphics[width=0.3\linewidth]{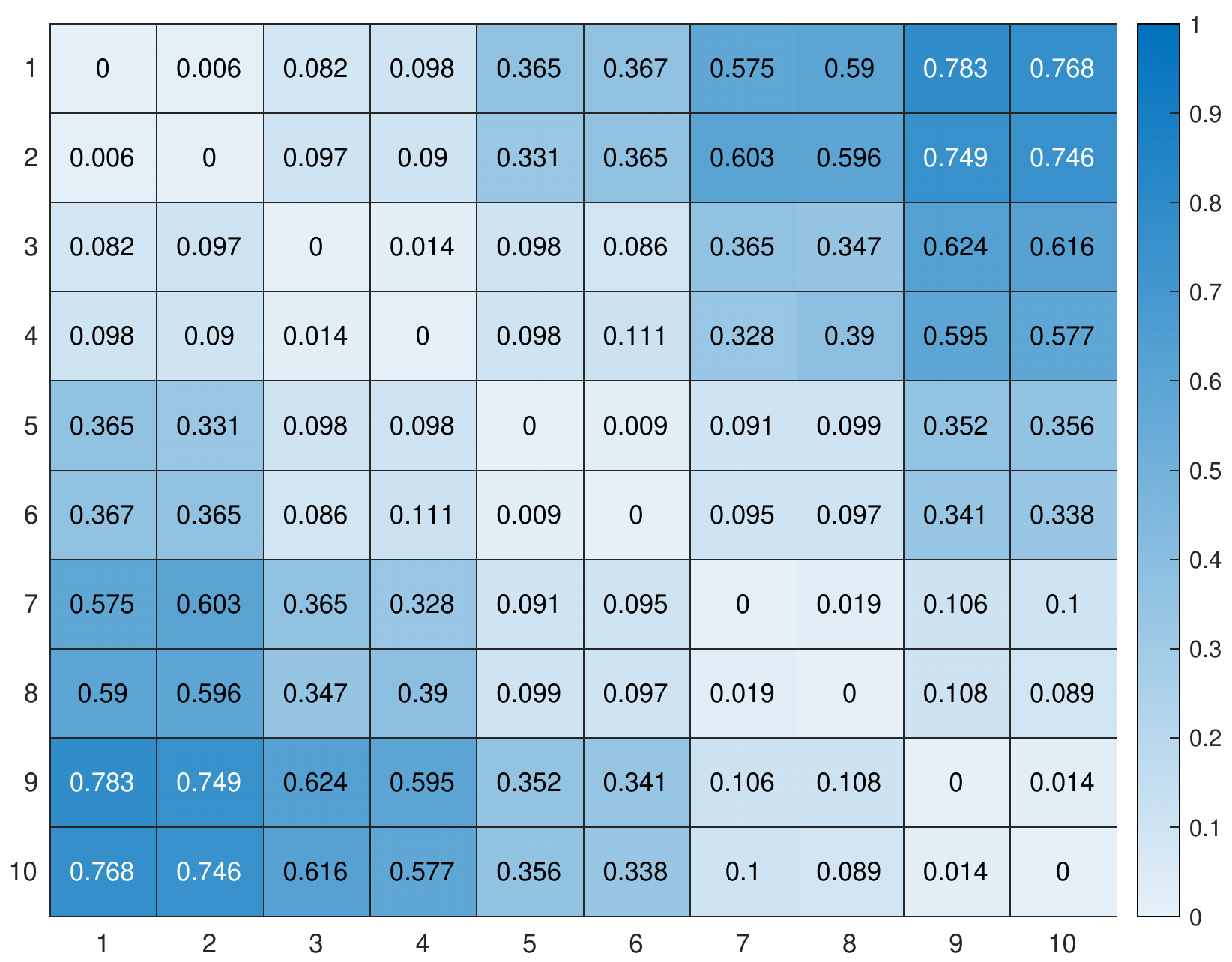}
}
\hfill
\subfigure[$\sigma=3$]{
\includegraphics[width=0.3\linewidth]{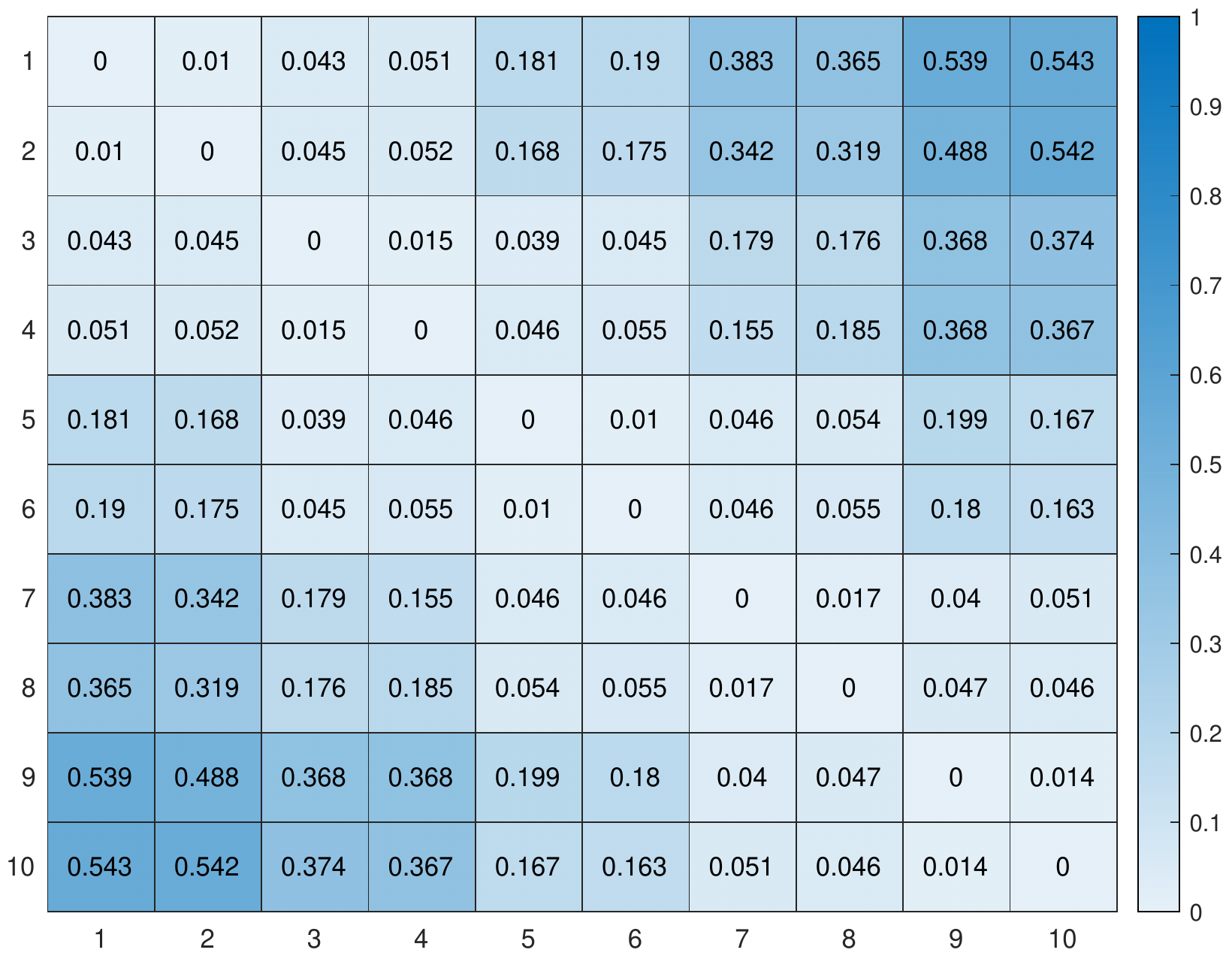}
}
\hspace{-2mm}

\vspace{-2mm}
\caption{Average rejection rate of the null hypothesis of \eqref{eq: null} for $\tau=0$ and  features with isotropic Gaussian distribution $x\sim \normal(0,I_{10})$. In this experiment, we consider $y|x\sim\normal(x^\sT \th^*,1)$ for $\th^*=[1,1,2,2,3,3,4,4,5,5]$. In this experiment the symmetric influence holds for pairs of features $(1,2),(3,4),(5,6),(7,8),$ and $(9,10)$. The small cell $(i,j)$ in each plot, represents rejection rates for testing symmetric influence for features $i$ and $j$ at significance level $\alpha=0.1$. In this experiment, the number of data points is $1000$ and the method is run with the score function $T(x,y)=|y-x^\sT \hth|$ for $\hth\sim\normal(\th^*,\sigma^2 I_{10})$ for three different $\sigma$ values $\sigma=1,2,3$ (from left to right). The reported numbers are averaged over $1000$ experiments.}
\label{fig: power-isotropic-appx}
\end{figure*}

We refer to Figure \ref{fig: power-isotropic-appx} for experiment on power of the test. 

\subsection{binary classification under mixture of Gaussians}
\begin{figure*}[t]
\centering
\subfigure[$n=5000$]{ 
\includegraphics[width=0.3\linewidth]{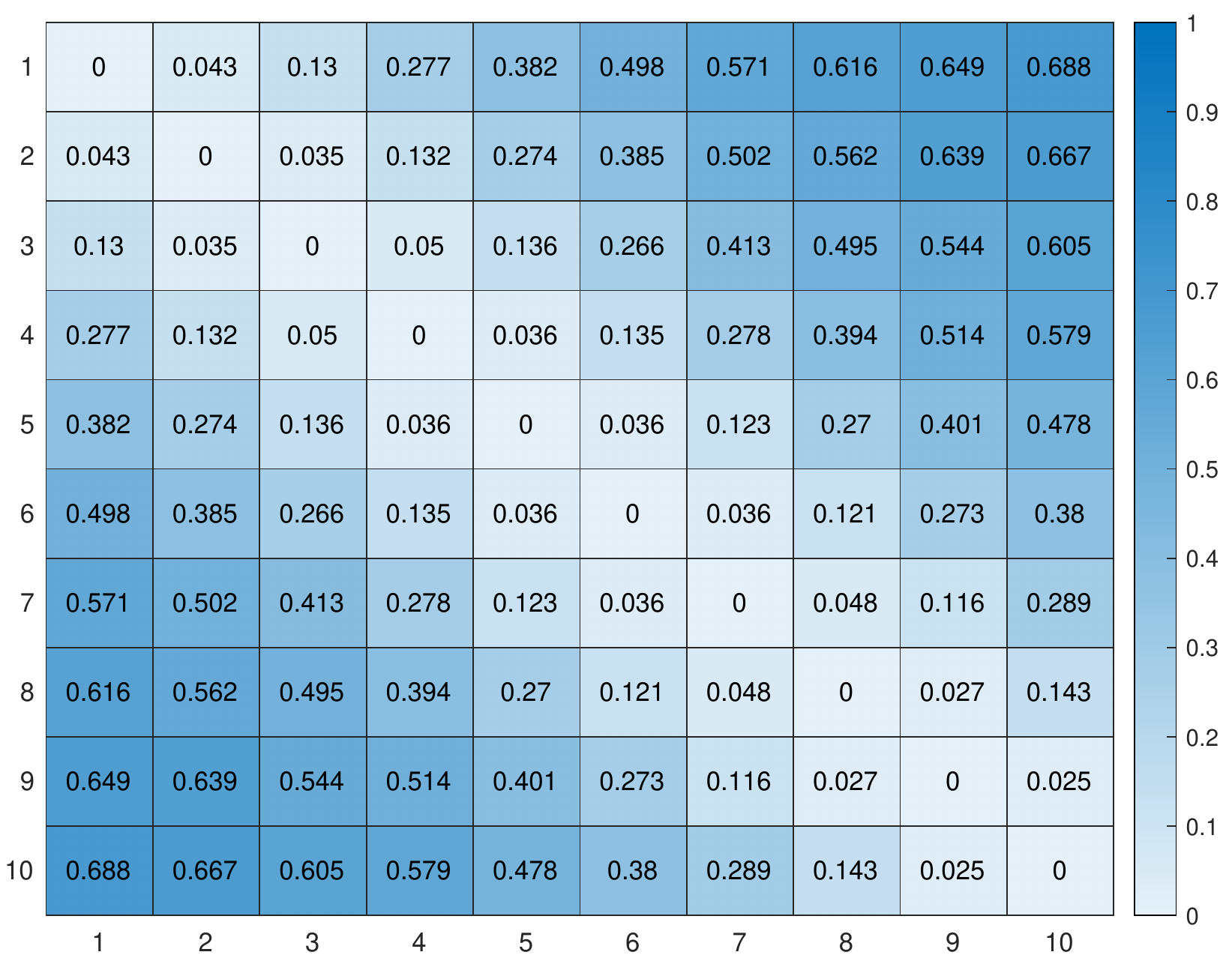}
}
\hfill
\subfigure[$n=20000$]{
\includegraphics[width=0.3\linewidth]{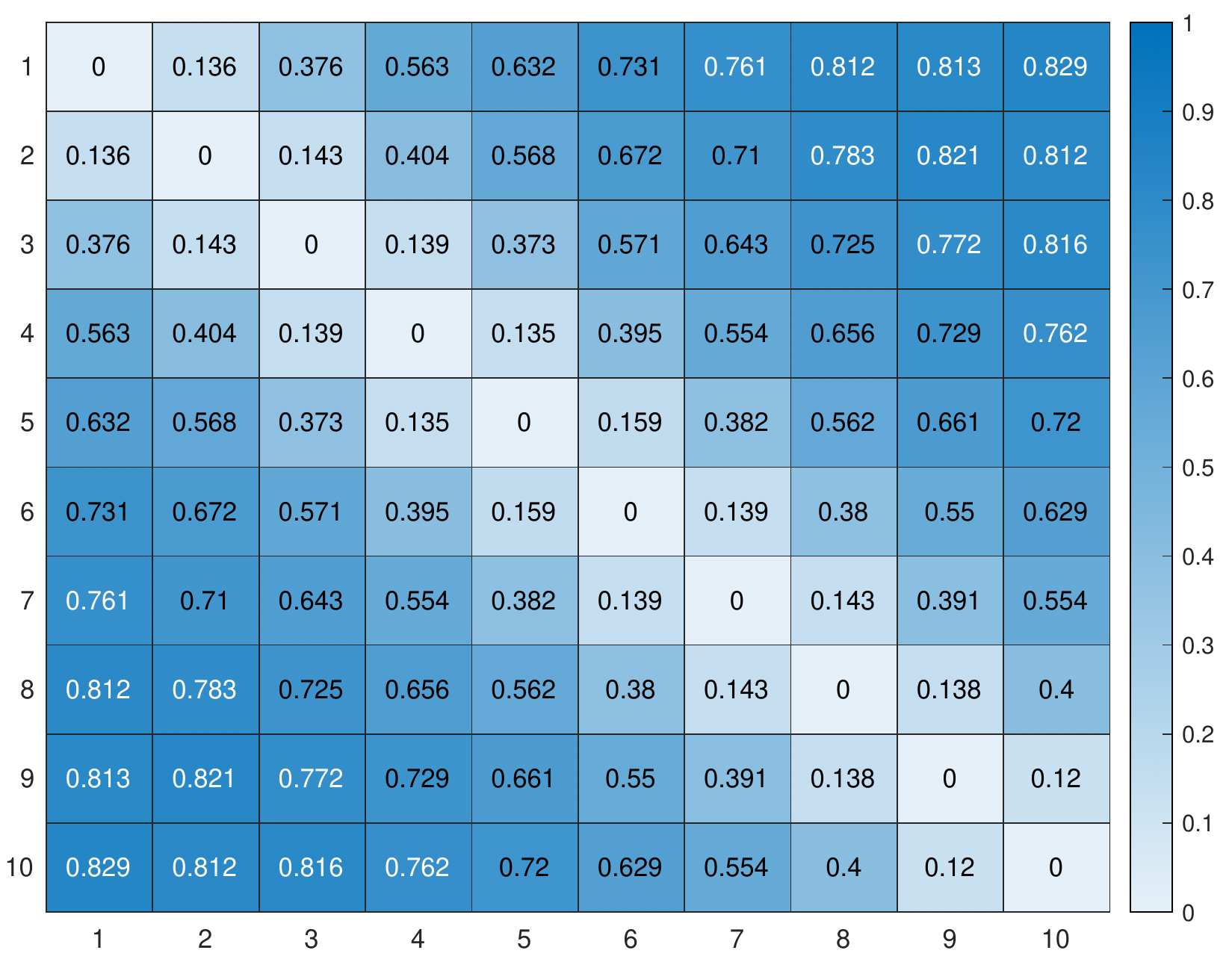}
}
\hfill
\subfigure[$n=50000$]{
\includegraphics[width=0.3\linewidth]{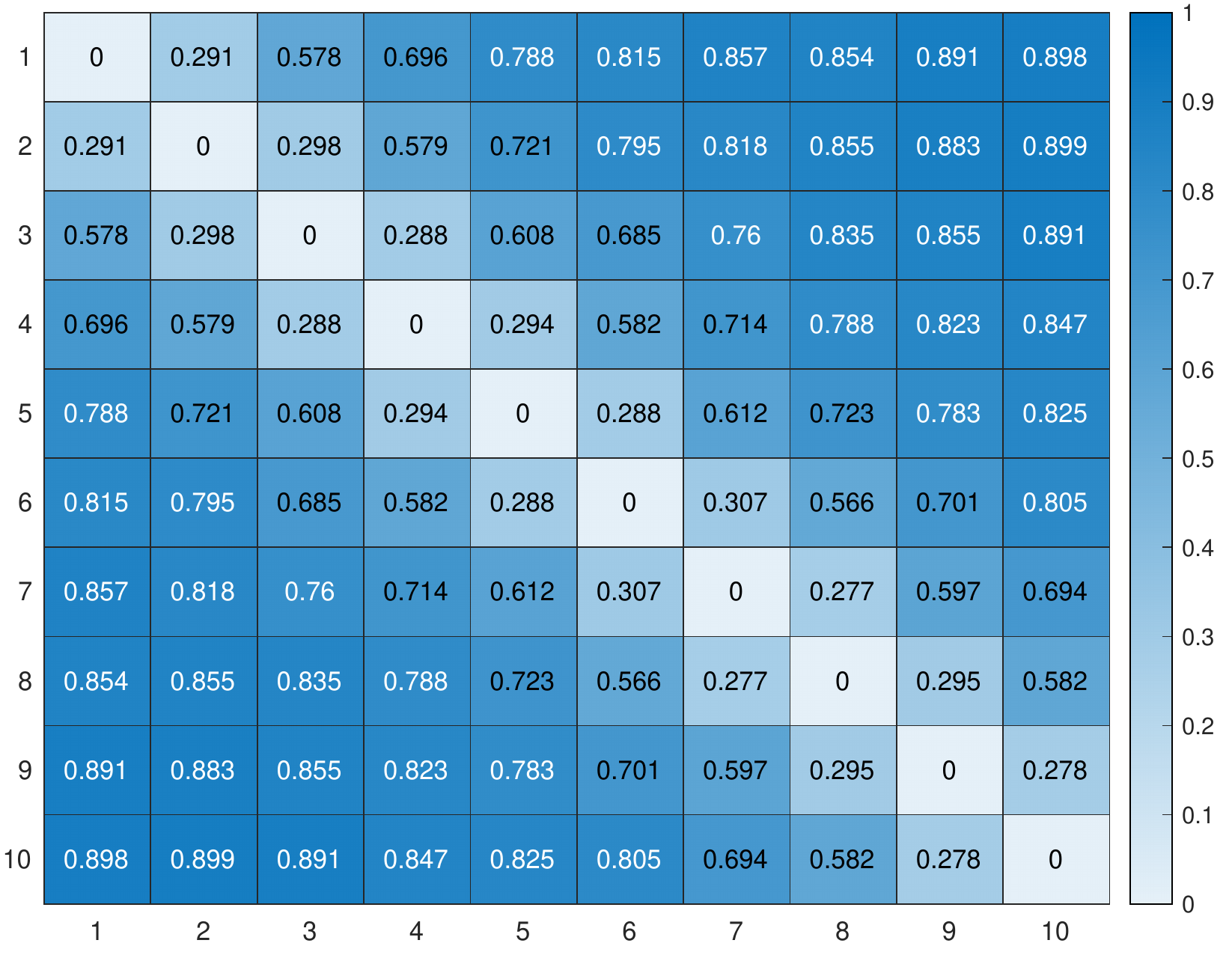}
}
\hspace{-2mm}

\vspace{-2mm}
\caption{Average rejection rate of the null hypothesis of \eqref{eq: null} for $\tau=0$ and  features with isotropic Gaussian distribution $x\sim \normal(0,I_{10})$ In this experiment, we consider binary classification under the mixture of Gaussian model \ref{eq: GMM} for $q=1/2$ and $\mu=\frac{\mu}{\|\tmu\|_2}$ for $\tmu=[1,2,\dots,10]$. The small cell $(i,j)$ in each plot, represents rejection rates for testing symmetric influence for features $i$ and $j$ at significance level $\alpha=0.1$. In this experiment, three different values for number of data points is considered $n=5000,20000, 50000$ (from left to right). We run Algorithm \ref{algorithm: importance} with the score function $T(x,y)=yx^\sT\hth$ for $\hth\sim\normal(0,I_{10})$. The reported numbers are averaged over $1000$ experiments.} \label{fig: binary}
\end{figure*}

In this section, we consider the problem of testing for symmetric influence for binary classification under a mixture of Gaussian model.  We consider the data generative law \eqref{eq: GMM} with $q=1/2$ and feature dimension $d=10$. We consider $\tmu=[1,2,3,\dots, 10]$ and let $\mu=\frac{\tmu}{\|\tmu\|_2}$. We follow the score function given in Theorem \ref{thm: power-binary} and consider $T(x,y)=y\hth^\sT x$ for some $\hth\sim \normal(0,I_d)$. We consider three different number of samples $n=5000, 20000, 50000$ for this experiment. Figure \ref{fig: binary} denote the results. Each number is averaged over $1000$ independent experiments. It can be observed that pairs with higher contrast between their $\mu$ values are rejected more often.

\subsection{robustness of data models experiment}
In the second experiment, we consider a pair of training samples with $5$ target examples. The first four targets are statistically significant (at level $\alpha=0.05$), while the target $5$ gives $pval=0.21$. We then replace the two training samples with some of their close other pictures, and compute the p-values for the new pair of images. We can see that the obtained p-values are somewhat close to the previous examples, which indicates the robustness of output results. The images along with p-values can be seen in Table 2. 
\begin{table*}[t]
\begin{center}
{\small
\begin{tabular}{|C C||C|C|C|C|C|}
\toprule
 Train pair& & Target 1 & Target 2 & Target 3&Target 4& Target 5 \\ 
\midrule
  \includegraphics[scale=1.3]{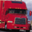}\caption{} & \includegraphics[scale=1.3]{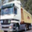}\caption{}  & \includegraphics[scale=1]{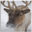}\caption{pval=$0.0023$}  & \includegraphics[scale=1]{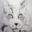}\caption{pval=$0.0026$} &
 \includegraphics[scale=1]{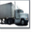}\caption{pval=$0.0092$}
 & \includegraphics[scale=1]{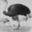}\caption{pval=$0.0173$}
&
 \includegraphics[scale=1]{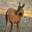}\caption{pval=$0.2108$}\\
 \bottomrule
   \includegraphics[scale=1.3]{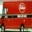}\caption{} & \includegraphics[scale=1.3]{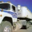}\caption{}  & \includegraphics[scale=1]{images_cifar/test_11.png}\caption{pval=$0.0023$}  & \includegraphics[scale=1]{images_cifar/test_12.png}\caption{pval=$0.0026$} &
 \includegraphics[scale=1]{images_cifar/test_13.png}\caption{{pval=${0.0084}$}}
 & \includegraphics[scale=1]{images_cifar/test_14.png}\caption{pval=$0.0188$}
&
 \includegraphics[scale=1]{images_cifar/test_15.png}\caption{pval=$0.2108$}\\
\midrule
\end{tabular} 
\caption{Table 2. Verifying the robustness of our findings for two pairs of training samples, that are highly close to each other.}
} 
\end{center}
\end{table*} 

%% file: main.bbl
\begin{thebibliography}{27}
\providecommand{\natexlab}[1]{#1}
\providecommand{\url}[1]{\texttt{#1}}
\expandafter\ifx\csname urlstyle\endcsname\relax
  \providecommand{\doi}[1]{doi: #1}\else
  \providecommand{\doi}{doi: \begingroup \urlstyle{rm}\Url}\fi

\bibitem[Bamber(1975)]{bamber1975area}
Bamber, D.
\newblock The area above the ordinal dominance graph and the area below the
  receiver operating characteristic graph.
\newblock \emph{Journal of mathematical psychology}, 12\penalty0 (4):\penalty0
  387--415, 1975.

\bibitem[Belloni et~al.(2014)Belloni, Chernozhukov, and
  Hansen]{belloni2014inference}
Belloni, A., Chernozhukov, V., and Hansen, C.
\newblock Inference on treatment effects after selection among high-dimensional
  controls.
\newblock \emph{The Review of Economic Studies}, 81\penalty0 (2):\penalty0
  608--650, 2014.

\bibitem[Benjamini \& Yekutieli(2001)Benjamini and
  Yekutieli]{benjamini2001control}
Benjamini, Y. and Yekutieli, D.
\newblock The control of the false discovery rate in multiple testing under
  dependency.
\newblock \emph{Annals of statistics}, pp.\  1165--1188, 2001.

\bibitem[Berrett et~al.(2020)Berrett, Wang, Barber, and
  Samworth]{berrett2020conditional}
Berrett, T.~B., Wang, Y., Barber, R.~F., and Samworth, R.~J.
\newblock The conditional permutation test for independence while controlling
  for confounders.
\newblock \emph{Journal of the Royal Statistical Society: Series B (Statistical
  Methodology)}, 82\penalty0 (1):\penalty0 175--197, 2020.

\bibitem[Bien et~al.(2021)Bien, Yan, Simpson, and M{\"u}ller]{bien2021tree}
Bien, J., Yan, X., Simpson, L., and M{\"u}ller, C.~L.
\newblock Tree-aggregated predictive modeling of microbiome data.
\newblock \emph{Scientific Reports}, 11\penalty0 (1):\penalty0 1--13, 2021.

\bibitem[Bogdan et~al.(2015)Bogdan, Van Den~Berg, Sabatti, Su, and
  Cand{\`e}s]{bogdan2015slope}
Bogdan, M., Van Den~Berg, E., Sabatti, C., Su, W., and Cand{\`e}s, E.~J.
\newblock Slope—adaptive variable selection via convex optimization.
\newblock \emph{The annals of applied statistics}, 9\penalty0 (3):\penalty0
  1103, 2015.

\bibitem[Candes \& Tao(2007)Candes and Tao]{candes2007dantzig}
Candes, E. and Tao, T.
\newblock The dantzig selector: Statistical estimation when p is much larger
  than n.
\newblock \emph{The annals of Statistics}, 35\penalty0 (6):\penalty0
  2313--2351, 2007.

\bibitem[Candes et~al.(2018)Candes, Fan, Janson, and Lv]{candes2018panning}
Candes, E., Fan, Y., Janson, L., and Lv, J.
\newblock Panning for gold:‘model-x’knockoffs for high dimensional
  controlled variable selection.
\newblock \emph{Journal of the Royal Statistical Society: Series B (Statistical
  Methodology)}, 80\penalty0 (3):\penalty0 551--577, 2018.

\bibitem[Cong et~al.(2013)Cong, Ran, Cox, Lin, Barretto, Habib, Hsu, Wu, Jiang,
  Marraffini, et~al.]{cong2013multiplex}
Cong, L., Ran, F.~A., Cox, D., Lin, S., Barretto, R., Habib, N., Hsu, P.~D.,
  Wu, X., Jiang, W., Marraffini, L.~A., et~al.
\newblock Multiplex genome engineering using crispr/cas systems.
\newblock \emph{Science}, 339\penalty0 (6121):\penalty0 819--823, 2013.

\bibitem[Crawford et~al.(2018)Crawford, Wood, Zhou, and
  Mukherjee]{crawford2018bayesian}
Crawford, L., Wood, K.~C., Zhou, X., and Mukherjee, S.
\newblock Bayesian approximate kernel regression with variable selection.
\newblock \emph{Journal of the American Statistical Association}, 113\penalty0
  (524):\penalty0 1710--1721, 2018.

\bibitem[Deshpande et~al.(2019)Deshpande, Javanmard, and
  Mehrabi]{deshpande2019online}
Deshpande, Y., Javanmard, A., and Mehrabi, M.
\newblock Online debiasing for adaptively collected high-dimensional data.
\newblock \emph{arXiv preprint arXiv:1911.01040}, 2019.

\bibitem[Duchi(2007)]{duchi2007derivations}
Duchi, J.
\newblock Derivations for linear algebra and optimization.
\newblock \emph{Berkeley, California}, 3\penalty0 (1):\penalty0 2325--5870,
  2007.

\bibitem[Fei \& Li(2021)Fei and Li]{fei2021estimation}
Fei, Z. and Li, Y.
\newblock Estimation and inference for high dimensional generalized linear
  models: A splitting and smoothing approach.
\newblock \emph{J. Mach. Learn. Res.}, 22:\penalty0 58--1, 2021.

\bibitem[Hsieh \& Turnbull(1996)Hsieh and Turnbull]{hsieh1996nonparametric}
Hsieh, F. and Turnbull, B.~W.
\newblock Nonparametric and semiparametric estimation of the receiver operating
  characteristic curve.
\newblock \emph{The annals of statistics}, 24\penalty0 (1):\penalty0 25--40,
  1996.

\bibitem[Ilyas et~al.(2022)Ilyas, Park, Engstrom, Leclerc, and
  Madry]{ilyas2022datamodels}
Ilyas, A., Park, S.~M., Engstrom, L., Leclerc, G., and Madry, A.
\newblock Datamodels: Predicting predictions from training data.
\newblock \emph{arXiv preprint arXiv:2202.00622}, 2022.

\bibitem[Javanmard \& Mehrabi(2021)Javanmard and Mehrabi]{javanmard2021pearson}
Javanmard, A. and Mehrabi, M.
\newblock Pearson chi-squared conditional randomization test.
\newblock \emph{arXiv preprint arXiv:2111.00027}, 2021.

\bibitem[Javanmard \& Montanari(2014)Javanmard and
  Montanari]{javanmard2014confidence}
Javanmard, A. and Montanari, A.
\newblock Confidence intervals and hypothesis testing for high-dimensional
  regression.
\newblock \emph{The Journal of Machine Learning Research}, 15\penalty0
  (1):\penalty0 2869--2909, 2014.

\bibitem[Krizhevsky et~al.(2009)Krizhevsky, Hinton,
  et~al.]{krizhevsky2009learning}
Krizhevsky, A., Hinton, G., et~al.
\newblock Learning multiple layers of features from tiny images.
\newblock 2009.

\bibitem[Liu et~al.(2022)Liu, Katsevich, Janson, and Ramdas]{liu2022fast}
Liu, M., Katsevich, E., Janson, L., and Ramdas, A.
\newblock Fast and powerful conditional randomization testing via distillation.
\newblock \emph{Biometrika}, 109\penalty0 (2):\penalty0 277--293, 2022.

\bibitem[Peters et~al.(2016)Peters, Colavin, Shi, Czarny, Larson, Wong,
  Hawkins, Lu, Koo, Marta, et~al.]{peters2016comprehensive}
Peters, J.~M., Colavin, A., Shi, H., Czarny, T.~L., Larson, M.~H., Wong, S.,
  Hawkins, J.~S., Lu, C.~H., Koo, B.-M., Marta, E., et~al.
\newblock A comprehensive, crispr-based functional analysis of essential genes
  in bacteria.
\newblock \emph{Cell}, 165\penalty0 (6):\penalty0 1493--1506, 2016.

\bibitem[Shaer \& Romano(2022)Shaer and Romano]{shaer2022learning}
Shaer, S. and Romano, Y.
\newblock Learning to increase the power of conditional randomization tests.
\newblock \emph{arXiv preprint arXiv:2207.01022}, 2022.

\bibitem[Shao et~al.(2021)Shao, Bien, and Javanmard]{shao2021controlling}
Shao, S., Bien, J., and Javanmard, A.
\newblock Controlling the false split rate in tree-based aggregation.
\newblock \emph{arXiv preprint arXiv:2108.05350}, 2021.

\bibitem[Tibshirani(1996)]{tibshirani1996regression}
Tibshirani, R.
\newblock Regression shrinkage and selection via the lasso.
\newblock \emph{Journal of the Royal Statistical Society: Series B
  (Methodological)}, 58\penalty0 (1):\penalty0 267--288, 1996.

\bibitem[Tibshirani et~al.(2005)Tibshirani, Saunders, Rosset, Zhu, and
  Knight]{tibshirani2005sparsity}
Tibshirani, R., Saunders, M., Rosset, S., Zhu, J., and Knight, K.
\newblock Sparsity and smoothness via the fused lasso.
\newblock \emph{Journal of the Royal Statistical Society: Series B (Statistical
  Methodology)}, 67\penalty0 (1):\penalty0 91--108, 2005.

\bibitem[Van~de Geer et~al.(2014)Van~de Geer, B{\"u}hlmann, Ritov, and
  Dezeure]{van2014asymptotically}
Van~de Geer, S., B{\"u}hlmann, P., Ritov, Y., and Dezeure, R.
\newblock On asymptotically optimal confidence regions and tests for
  high-dimensional models.
\newblock \emph{The Annals of Statistics}, 42\penalty0 (3):\penalty0
  1166--1202, 2014.

\bibitem[Wilms \& Bien(2022)Wilms and Bien]{wilms2022tree}
Wilms, I. and Bien, J.
\newblock Tree-based node aggregation in sparse graphical models.
\newblock \emph{Journal of Machine Learning Research}, 23\penalty0
  (243):\penalty0 1--36, 2022.

\bibitem[Yan \& Bien(2021)Yan and Bien]{yan2021rare}
Yan, X. and Bien, J.
\newblock Rare feature selection in high dimensions.
\newblock \emph{Journal of the American Statistical Association}, 116\penalty0
  (534):\penalty0 887--900, 2021.

\end{thebibliography}
